\documentclass[10pt,twocolumn,letterpaper]{article}

\usepackage{wacv}              

\usepackage{graphicx}
\usepackage{amsmath}
\usepackage{amssymb}
\usepackage{booktabs}
\usepackage{pifont}
\usepackage[]{prettyref}
\usepackage{steinmetz} 
\usepackage{caption}
\usepackage{subcaption}
\usepackage{float}
\usepackage{amsmath,amsfonts}
\usepackage{mathrsfs}
\usepackage[flushleft]{threeparttable}
\usepackage{rotating}
\usepackage{lscape}
\usepackage{amsmath,amssymb,amsfonts}
\usepackage{breqn}
\usepackage{datatool}
\usepackage{pifont}
\usepackage{setspace}
\usepackage{arydshln}
\usepackage{tikz}
\usepackage{multirow}
\usepackage{algorithm}
\usepackage{algpseudocode}
\usepackage{placeins}

\usepackage{slashbox}

\usepackage[accsupp]{axessibility}  

\usepackage[pagebackref,breaklinks,colorlinks]{hyperref}

\usepackage[capitalize]{cleveref}
\crefname{section}{Sec.}{Secs.}
\Crefname{section}{Section}{Sections}
\Crefname{table}{Table}{Tables}
\crefname{table}{Tab.}{Tabs.}

\def\wacvPaperID{2760} 
\def\confName{WACV}
\def\confYear{2025}

\begin{document}

\title{Autoregressive Adaptive Hypergraph Transformer for \\ Skeleton-based Activity Recognition}

\author{Abhisek Ray$^{1}$  Ayush Raj$^{1}$  Maheshkumar H. Kolekar$^{1}$ \\
	$^{1}$Indian Institute of Technology Patna, India \\
	{\tt\small rayabhisek0610@gmail.com, \{ayush_2001ee10, mahesh\}@iitp.ac.in}
}

\maketitle

\begin{abstract}
     Extracting multiscale contextual information and higher-order correlations among skeleton sequences using Graph Convolutional Networks (GCNs) alone is inadequate for effective action classification. Hypergraph convolution addresses the above issues but cannot harness the long-range dependencies. The transformer proves to be effective in capturing these dependencies and making complex contextual features accessible. We propose an \textbf{Autoreg}ressive \textbf{Ad}aptive \textbf{H}yper\textbf{G}raph Trans\textbf{former} (\textbf{AutoregAd-HGformer}) model for in-phase (autoregressive and discrete) and out-phase (adaptive) hypergraph generation. The vector quantized in-phase hypergraph equipped with powerful autoregressive learned priors produces a more robust and informative representation suitable for hyperedge formation. The out-phase hypergraph generator provides a model-agnostic hyperedge learning technique to align the attributes with input skeleton embedding. The hybrid (supervised and unsupervised) learning in AutoregAd-HGformer explores the action-dependent feature along spatial, temporal, and channel dimensions. The extensive experimental results and ablation study indicate the superiority of our model over state-of-the-art hypergraph architectures on the NTU RGB+D, NTU RGB+D 120, and NW-UCLA datasets. Find the code \href{https://github.com/rayabhisek123/AutoregAd-HGformer}{Here}.
\end{abstract}
\vspace*{-4mm}

\section{Introduction}
\label{sec:intro}
\vspace*{-2mm} 
In recognizing various human actions, less attention has been paid to the body skeleton compared to other modalities \cite{wang2015action, tran2018closer, zhao2017temporal} like appearance, optical flow, and depth \cite{yan2018spatial}. Nowadays, skeleton-based action recognition has gained popularity due to its immutable nature towards view-point variation, illumination changes, and background clutter. Reduced computational cost and background adaptation widen its application in surveillance, smart security, healthcare, and Human-Computer Interaction (HCI) \cite{wang2019comparative, xin2023transformer}.

Earlier, the non-euclidean graph geometry of skeleton sequences related to human actions is analyzed through various fully supervised architectures such as image-based Convolutional Neural Networks (CNNs) \cite{ke2017new, soo2017interpretable, li2017skeleton}, sequence-based Recurrent Neural Networks (RNNs) \cite{liu2016spatio, zhu2016co, shahroudy2016ntu, zhang2017geometric}, graph-based Graph Convolutional Networks (GNNs) \cite{yan2018spatial, chi2022infogcn, chen2021channel, lee2023hierarchically, song2022constructing}, and attention-based transformers \cite{pang2022igformer, long2023step, gao2022focal, wu2023skeletonmae, mao2023masked}. Among them, GCN-based architecture is widely adopted because the inductive biases present in CNNs or RNNs are unable to capture the non-euclidean joint-bone attributes of the human skeleton. 

The ST-GCN \cite{yan2018spatial}, a pioneering architecture for graph geometry, uses a fixed sampling method for selecting the neighborhood to perform convolution operations on skeleton sequences. However, the predefined adjacency matrix, which represents the natural connections of a human skeleton, fails to address the high-order correlations between the farthest skeleton nodes and is unable to capture the multiscale semantic information. To get rid of the above drawbacks, adaptive graph topologies are introduced, combining predefined adjacency graph topology with either learnable node topology \cite{shi2019two} or learnable channel topology \cite{chen2021channel} to fetch action-dependent skeleton features. The above approaches manage to aggregate these features to alleviate the performance, but the scale of effectiveness still needs to be higher to handle the diversity in state-of-the-art datasets. While several architectures \cite{lee2023hierarchically,wei2023accommodating,chi2022infogcn,chen2021channel} have emerged to address these challenges, their effectiveness has been only partially realized.

\textbf{Hypergraphs over graph topology:} Hypergraphs \cite{10220028, zhu2022selective, hao2021hypergraph} offer significant advantages over traditional graphs in skeleton-based activity recognition by capturing higher-order relationships among joints, which are crucial for actions like ``waving," where multiple joints (shoulder, elbow, wrist) need to be considered simultaneously. They represent multi-scale contextual information, allowing for a detailed understanding of local and global patterns in actions such as ``running" or ``walking," where limb and full-body coordination are both essential. Hypergraphs are also more robust to noise by emphasizing critical joints, which helps in actions like ``sitting down," where irrelevant joint movements can mislead simpler models. Enhanced feature fusion in hypergraphs aids in recognizing actions like ``clapping" by effectively combining spatial and temporal dynamics across multiple joints. They also adapt dynamically, adjusting joint importance throughout actions like ``jumping," where the focus shifts between legs and arms during the sequence. In complex activities such as ``picking up and throwing an object," hypergraphs excel by capturing the intricate interdependencies among multiple body parts involved in the motion. These capabilities make hypergraphs particularly powerful for accurately recognizing a wide range of skeleton-based actions.

\begin{figure*}[h!]
    \centering
        \vspace*{-6mm}
    \captionsetup{justification=centering}
    \includegraphics[width=0.975\textwidth]{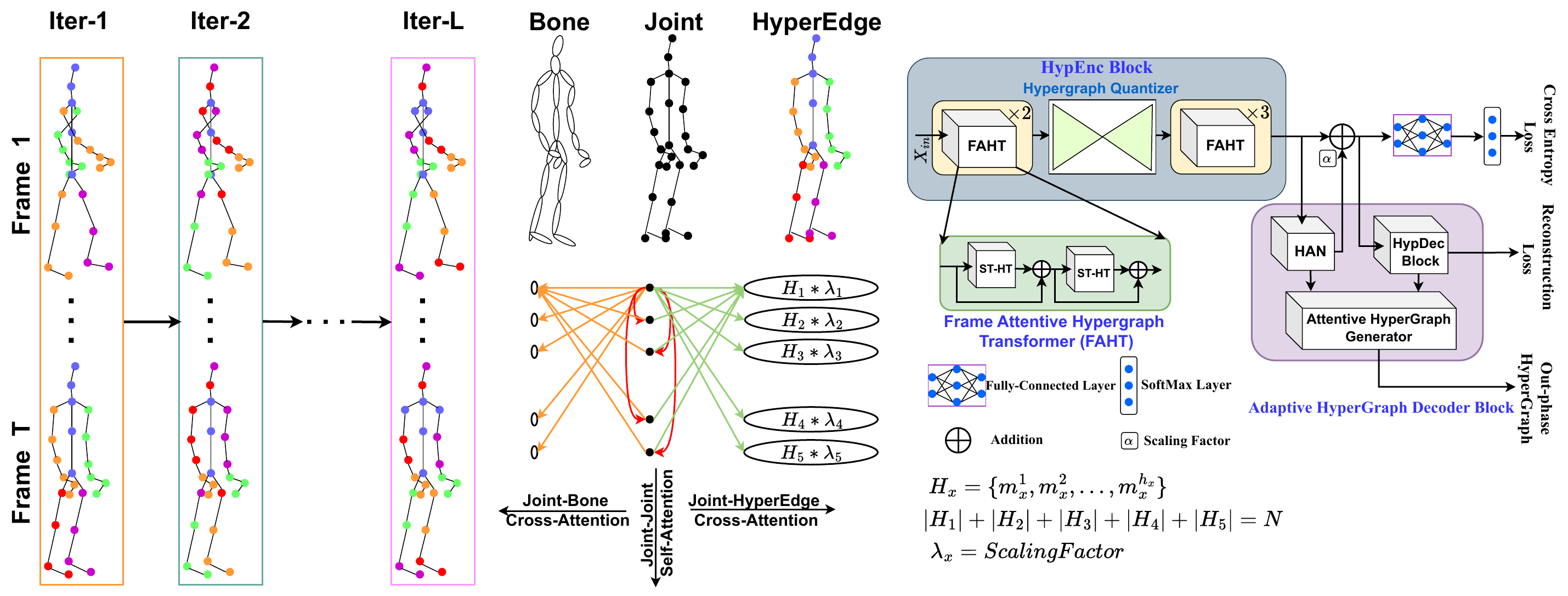}
    \vspace*{-2mm} 
    \caption{Model abstraction. Model-agnostic iterative hypergraph (left), various attention (middle) and AutoregAd-HGformer (right)}
    \label{fig:WACV_Model}
    \vspace*{-6mm}
\end{figure*}

Two primary strategies for generating hyperedges to form a hypergraph from skeletal sequences are: (i) fixed-generated hypergraph \cite{zhou2022hypergraph} and (ii) dynamically generated hypergraph \cite{wei2021dynamic, hao2021hypergraph, wang2023dynamic} techniques. Since the set of nodes in each hyperedge is defined by one-to-one correlations between body joints within a specific action frame, a fixed hypergraph cannot effectively capture the dynamic, action-dependent contextual features. While dynamically generated hypergraphs can partially incorporate action-dependent attributes \cite{wei2021dynamic, wang2023dynamic}, the frequent transitions of the hypergraph inside the descriptors disrupt the feature distribution, thereby limiting its effectiveness for classification. However, we propose in-phase and out-phase hypergraph techniques to boost representation for intra-batch and inter-batch embeddings. By applying vector quantization with autoregressive learned priors, the input hypergraph is discretized to produce the in-phase hypergraph inside the feature extractor (encoder). It prevents frequent node transition between hyperedges and boosts robustness and dynamic adaptation by providing better representation. Similarly, we also designed a decoder that generates model-agnostic out-phase hypergraphs outside the extractor. Here, the hypergraph is adaptively restructured in response to action-specific features. It takes hypergraph-attentive spatiotemporal features from the encoder to generate an action-dependent model-independent hypergraph for the next iteration. 

We adopt a unique transformer design that analyzes individual features of joint and hyperedge along with their mutual semantics. The overall abstraction behind our model implementation is shown in Fig. \ref{fig:WACV_Model}. The contributions of the proposed AutoregAd-HGformer can be summarized as:
\begin{enumerate}
    \item We propose an autoregressive adaptive HyperGraph Transformer (AutoregAd-HGformer) model that introduces a unique transformer-implemented hypergraph architecture to skeleton-based action sequences and mutates hyperedge configuration adaptively to avail more diverse features for recognition.
    
    \item We propose two unique hypergraph generation techniques to produce in-phase and out-phase hypergraphs for discrete and continuous feature alignment, respectively. Leveraging both model-dependent and model-agnostic hypergraphs, AutoregAd-HGformer preserves intra-batch and inter-batch relationships, thereby enhancing the richness of hypergraph representations.
    
    \item The hybrid learning (supervised and self-supervised) in AutoregAd-HGformer explores the action-dependent feature along spatial, temporal, and channel dimensions.
    
    \item The extensive experimental results and ablation study indicate the superiority of our model over state-of-the-art hypergraph architectures on NTU RGB+D, NTU RGB+D 120, and NW-UCLA datasets.
\end{enumerate}
\vspace*{-1mm}

\section{Related Works}
\label{sec:Related_Works}
\vspace{-1mm}
Numerous graph architectures, including graph convolution networks, auto-encoders, transformers, and hypergraph convolution networks, address non-Euclidean data geometry. Our hypergraph framework leverages multi-scale semantics and non-linear contextual features using hypergraph convolution and transformers. These methods play a crucial role in developing the novel hypergraph transformer-based AutoregAd-HGformer.
\vspace{-1mm}

\subsection{Graph Transformer Methods}
\label{sec:Transformer_Methods}
\vspace{-1mm}
To move beyond handcrafted traversal rules or graph topologies, DSTA-Net introduces a transformer-based architecture using decoupled spatiotemporal self-attention for skeleton-based action and gesture recognition \cite{shi2020decoupled}. Depending on the sampling rate, the temporal stream is split into fast and slow streams. SAN \cite{cho2020self} presents three network variants using multi-head self-attention on encoded or fused skeleton features of action sequences. Plizzari et al. \cite{plizzari2021spatial} developed a spatial-temporal transformer for skeleton-based activity recognition, learning intra-frame interaction and inter-frame correlation through spatial and temporal convolutions and transformers. STSA \cite{aksan2021spatio} decouples spatial and temporal self-attention in parallel with a combined feed-forward network. Zhang et al. \cite{zhang2021stst} designed STST, a sequential block of special and directional temporal transformers, for modeling skeleton-based action sequences. FG-STformer \cite{gao2022focal} emphasizes local joint and global body part features via self- and cross-attention layers. For skeleton-based human interaction recognition, \cite{pang2022igformer} proposes a semantic partition module generating Body-Part-Time sequences fed to IGFormer, which enhances graph representation with cross-attention. Hue et al. \cite{long2023step} apply a similar approach by dividing skeleton action sequences into upper-lower and hand-foot joints. All frameworks above use supervised learning with labeled skeletal action sequences. The scarcity of labeled data in human action recognition limits the full potential of transformers. Recently, self-supervised transformer architectures \cite{wu2023skeletonmae, mao2023masked} have been introduced, using masking techniques to improve performance. Wu et al. \cite{wu2023skeletonmae} proposed SkeletonMAE, employing a masking and reconstruction pipeline with a self-supervised encoder-decoder transformer. MAMP \cite{mao2023masked} introduces motion-aware masking based on motion intensity for motion predictions of joint embeddings from skeletal action sequences, utilizing multi-head self-attention in the encoder and decoder.
\vspace{-1mm}

\subsection{Hypergraph Convolution Methods}
\label{sec:Hypergraph_Convolution_Methods}
\vspace{-2mm}
Wei et al. introduced DHGCN, a hypergraph convolution network for skeleton-based action recognition using dynamic topology and dynamic joint weights \cite{wei2021dynamic}. Dynamic topology groups nodes based on relative features, and dynamic joint weights assign importance-based weights to nodes. However, the hypergraph's dynamic variation in each extraction layer can disrupt the deep features fed to the classification head. Concurrently, Hao et al. proposed Hyper-GNN, a hypergraph skeleton model that captures multiscale spatiotemporal intelligence and higher-order dependencies among skeleton sequences \cite{hao2021hypergraph}. While it reduces noise from uncorrelated joints, it does not address the limitations of \cite{wei2021dynamic}. In both \cite{wei2021dynamic} and \cite{hao2021hypergraph}, hyperedge construction is based on the moving distance between consecutive frames. Selective-HCN \cite{zhu2022selective} uses selective-scale hypergraph convolution (SHC) and selective-frame temporal convolution (STC) to aggregate spatiotemporal features. SHC captures and fuses multiscale information selectively, while STC extracts keyframes for temporal features. However, Selective-HCN cannot extract action correspondence features due to a fixed hypergraph and struggles with distinguishing similar actions at different paces. To address separate learning of spatiotemporal information, DST-HCN \cite{wang2023dynamic} dynamically and parallelly learns using dynamic hypergraph convolution and an information fusion block, but it still exhibits drawbacks similar to \cite{wei2021dynamic} and \cite{hao2021hypergraph}.

The AutoregAd-HGformer uses hypergraph convolution and the hypergraph transformer along with in-phase and out-phase hypergraph generation in the feature extraction unit to mitigate the above issues and effectively derive multiscale semantics and efficient motion features.
\vspace{-1mm}
\begin{figure*}[h!]
	\centering
	\captionsetup{justification=centering}
	\includegraphics[width=1.0\textwidth]{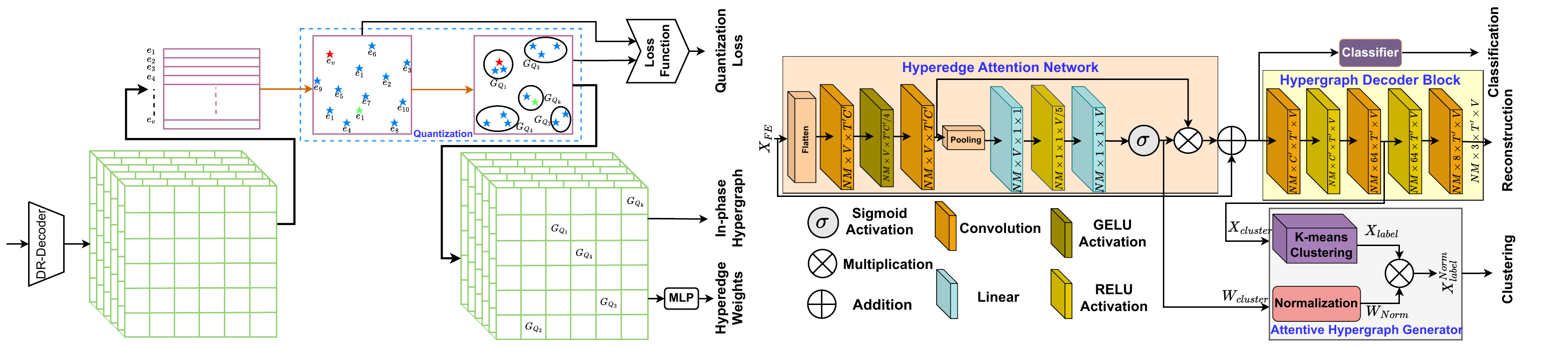}
	\vspace*{-4mm} 
	\caption{Proposed framework for autoregressive in-phase hypergraph quantizer (left) and adaptive hypergraph decoder (right).}
	\label{fig:Architecture_Ad-HGformer}
	\vspace*{-6mm} 
\end{figure*}

\section{Proposed Method}
\label{sec:Proposed_Method}
\vspace{-2mm}
As shown in Fig. \ref{fig:WACV_Model}, the proposed AutoregAd-HGformer comprises three functional blocks: (i) hypergraph encoder, (ii) adaptive hypergraph decoder, and (iii) classifier. A fully connected and a softmax layer are combined to form the classifier block. The other two blocks are discussed below:
\vspace{-1mm}

\subsection{Hypergraph Encoder}
\label{sec:Hypergraph_Encoder}
\vspace{-1mm}
The Hypergraph Encoder (HypEnc) block comprises a stack of five Frame Attentive Hypergraph Transformer (FAHT) units divided between two groups in a 2:3 ratio. The output embedding of the first group is passed through a hypergraph quantizer to find an autoregressive discrete hypergraph for the second group. The output embedding $E$ for each node of the HypEnc block  can be expressed as:  
\vspace{-2mm}
\begin{equation}
    E=\text{HypEnc}(\mathbf{A},X,\mathbf{H^n},\mathbf{{H_{w}}^n}),
    \vspace{-2mm}
\end{equation}

\noindent where \( \mathbf{A} \in \{0,1\}^{V \times V} \) is the adjacency matrix, \( \mathbf{X} \in \mathbb{R}^{N \times M \times V \times T \times C} \) represents the input feature embedding, \( \mathbf{H}^n \in \{0,1\}^{V \times E_h} \) is the incidence matrix, \( \mathbf{H_{w}}^n \in \mathbb{R}^{E_h \times E_h} \) denotes the hyperedge weight matrix, and \( \mathbf{E} \in \mathbb{R}^{NM \times V \times T' \times C'} \) corresponds to the output embedding. The parameters \( n \), \( N \), \( M \), \( V \), \( T \), \( C \), \( E_h \), \( T' \), and \( C' \) refer to the iteration number, batch size, number of individuals per frame, skeletal joint count, frame count per action, input channel count, number of hyperedges, number of output frames, and output channel count, respectively. Following in-phase and out-phase hyperedge calculations, the updated hypergraph is propagated to the next iteration every time. The FAHT unit in each group comprises the Spatio-Temporal Hypergraph Transformer (ST-HT) unit and the Spatiotemporal Attentive Hypergraph Transformer (STA-HT) unit.

\noindent{\textbf{ST-HT:}} The \textbf{H}yper\textbf{G}raph \textbf{C}onvolution ($HG_{C}$) is applied to find hypergraph embedding that subsequently passes to the transformer unit to calculate cross-attention. The hypergraph convolution is defined by 
\vspace{-2mm}
\begin{equation}
    \label{eq:HG_feature}
    H_f=\lambda_{1} \text{HG}_{C}(X,\mathbf{H^n},\mathbf{H_{w}^n})+\lambda_{2} \text{HG}_{C}(X,\mathbf{H_q^n},\mathbf{H^n_{q_w}}).
    \vspace{-1.5mm}
\end{equation}

\noindent \( \mathbf{Hq^n} \in \{0,1\}^{N*M \times V \times E_h} \) and \(\mathbf{H^n_{q_w}} \in \mathbb{R}^{N*M \times E_h \times E_h} \) are the quantized in-phase hypergraph and its weight, respectively. $H_n$ and $H^n_w$ are the out-phase hypergraph and its weight from the previous iteration, respectively. For the first two FAFT units, $\mathbf{H^n_q}=\mathbf{H^n}$ and $\mathbf{H^n_{q_w}}=\mathbf{H_{w}^n}$ as quantized hypergraph is calculated after the second unit as shown in Fig. \ref{fig:Architecture_Ad-HGformer}. The equations
\vspace{-2mm}
\begin{equation}
    {sa}^{l}_{i,j}=q^{l}_{i} {k^{l}_{j}}^T,
    {{ca}_h}^{l}_{i,j}=q^{l}_{i} {{H^{l}_{f}}_j}^T, {{ca}_r}^{l}_{i,j}=q^{l}_{i} {{P^{l}_{i,j}}}^T
    \vspace{-2mm}
\end{equation}
 represent the Joint-to-Joint self-attention, Joint-to-Hyperedge cross-attention and Joint-to-Bone cross-attention for the $l^{th}$ head, respectively. The $q$ and $k$ are linearly transformed query and key vectors. \( P^{l} \in \mathbb{R}^{N \times N \times ( C'/h)} \) is the bone attributes of the skeleton sequences. All these attention embeddings are aggregated and multiplied with value ($v$) vector by the following equation: 
 \vspace{-4mm}
\begin{equation}
	Y^l_{i} = \sum_{j=1}^{N} \frac{{a^{l}_t}_{i,j}}{\sum_{j} {a^{l}_t}_{i,j}}{v^{l}_{j}}.
	\vspace{-2mm}
\end{equation}
Here $a^l_{t_{i,j}}$ is the aggregated representation of $l^{th}$ head. At the end, all the heads \{$Y^{0}_i,Y^{1}_i,...,Y^{h-1}_i$\} are concatenated to form the final output $Y_i$ for the $ith$ node.

\noindent{\textbf{STA-HT:} All the functional units in STA-HT are the same as the ST-HT block, except temporal attention is applied to hypergraph features in channel dimensions to recognize the importance of each frame. Details can be found in the supplementary material.

\noindent{\textbf{Hypergraph Quantizer:}} As shown in Fig. \ref{fig:Architecture_Ad-HGformer}, we follow the autoregressive vector quantization method \cite{van2017neural} for discrete in-phase hypergraph generation. To escape from the curse of dimensionality \cite{chandra2023escaping}, the output embeddings \( {f} \in \mathbb{R}^{ V  \times D} \) from the second FAHT unit is passed through a decoder block to get an intermediate low-dimensional vector \( {E} \in \mathbb{R}^{ V  \times d} \). The decoder used here is very similar to the decoder used in the adaptive hypergraph decoder block (sec. \ref{sec:Adaptive_Hypergraph_Decoder}). The embedding E is passed through a discretization bottleneck and subsequently maps to the nearest hyperedge $Q_j$ with the help of autoregressive trainable vectors, which served as codebook \{$ G_{Q_1},...., G_{Q_K}$\}. The assignment operator for these discrete input-independent hyperedges is defined by
\vspace{-2mm}
\begin{equation}
    he_i = argmin_{j} \lVert E_i-Q_j \rVert_2,
    \vspace{-2mm}
\end{equation}
where \(|he_i|=K\) denotes the hyperedge. The quantized vector $Q_j$, also referred to as an in-phase hypergraph $H_q$, is passed through an MLP layer to get the weights of the hypergraph $H_{q_w}$. The objective of learning these vectors is similar to clustering the nodes during hypergraph formation. As quantization is a lossy and non-differentiable process, the gradient is calculated over the quantizer input 
instead of intermediate discrete embedding \cite{bengio2013estimating}.  
\vspace*{-1mm}

\subsection{Adaptive Hypergraph Decoder}
\label{sec:Adaptive_Hypergraph_Decoder}
\vspace*{-2mm} 
The features $E_{enc}$ from the hypergraph encoder block are brought for classification and out-phase hypergraph generation after passing through a Hyperedge Attention Network (HAN) as shown in Fig. \ref{fig:Architecture_Ad-HGformer}. The output embeddings $ A_t \in \mathbb{R}^{NM\times V \times T' \times C'} $ from HAN are passed through the decoder to reconstruct the skeleton sequences and provide low-dimensional features ${E_c} \in \mathbb{R}^{NM\times V  \times d}$ to the adaptive hypergraph generator, where the out-phase hypergraph is prompted. We take $d \ll C'$ due to the ``curse of dimensionality" during clustering. We take $d$ (hyperparameter) as 8 for a $C'$ of 128. The output from the second layer of the decoder is considered to calculate the model-agnostic out-phase hypergraph. The following equations describe the above process. 
\vspace{-2mm}
\begin{equation}
        \label{eq:HAN}
	A_t=\text{HAN}(E_{enc})
 \vspace{-2mm}
\end{equation}
\vspace{-4mm}
\begin{equation}
        \label{eq:HypDec}
	E_c = \text{HypDec}_{\text{s}}(\text{GAP}_{\text{time}}(E_{enc}+ \alpha A_t),\mathbf{A})
 \vspace{-2mm}
\end{equation}
\vspace{-4mm}
\begin{equation}
        \label{eq:Embc}
	Emb_c = \text{Norm}({\text{GAP}}_{\text{batch}}(E_c))
 \vspace{-2mm}
\end{equation}

\noindent Here, $\text{GAP}_{\text{time}}$, and $\text{GAP}_{\text{batch}}$ represent the normalization, global average pooling in the temporal dimension, and global average pooling across the total number of skeletons, respectively. 

\noindent{\textbf{Hyperedge Attention Network (HAN):}} 
For hyperedge attention, we compute the weights from the output embeddings of the HypEnc block using a squeeze-excitation network \cite{chen2022activating}.
\vspace{-2mm}
\begin{equation}
    E_2 = \text{Conv}_1(\text{GELU}(\text{Conv}_2(E_{enc})))
    \vspace{-2mm}
\end{equation}
We use dot hypergraph attention as shown in the equation:
\vspace{-2mm}
\begin{equation}
    A_t=attn \odot E_{2}, \text{where}
    \vspace{-2mm}
\end{equation}
\vspace{-9mm}

\begin{align}
    attn=&\sigma(F_{Lin_2}^{node}(\text{ReLU}(F_{Lin_1}^{node}(E_3)))), \\
    E_3=&\text{GAP}_{T'C'}(E_2).
    \vspace{-3mm}
\end{align}

\noindent Here, $\text{GAP}_{T'C'}$, $F_{Lin_1}^{node}$, $F_{Lin_2}^{node}$, and $\sigma$ represent the global average pooling along time and channel, two linear layers, and sigmoid activation layers, respectively.

\noindent{\textbf{Hypergraph Decoder (HypDec):}} The combined residual and attentive features from HAN are passed through the 5-layered decoders, where each layer performs the following hypergraph convolution.
\vspace{-2mm}
\begin{equation}
	\mathbf{Y} = \mathbf{D}^{-1/2}\mathbf{A_1}\mathbf{D}^{-1/2}\mathbf{X_1}\mathbf{W}.
    \vspace{-2mm}
\end{equation}
\noindent Here, $\mathbf{X_1} \in \mathbb{R}^{V_1\times C_1}$, $\mathbf{A_1} \in \{0, 1\}^{V \times V}$, D, and $\mathbf{W} \in \mathbb{R}^{C_1\times C_2}$ represent the decoder input, adjacency matrix, degree matrix, and weight matrix of the hypergraph, respectively. The output from the third layer is taken for out-phase hypergraph generation. The output feature from the decoder is used for graph reconstruction, which advocates self-supervised learning.

\noindent{\textbf{Attentive Hypergraph Generator:}} K-means clustering is applied to the intermediate output of the decoder to find hyperedges. The weights from HAN are normalized to calculate the importance of each hyperedge. The model-agnostic hypergraph can be expressed by 
\begin{equation}
\mathbf{H^{l+1}_{i,j}} =
\begin{cases}
	1 & \text{if joint \(i\) is assigned to cluster \(j\)} \\
	0 & \text{    otherwise}
\end{cases}
\vspace{-1mm}
\end{equation}
\noindent and their weights are defined by
\vspace{-1mm}
\begin{equation}
    \mathbf{H_{w}^{l+1}}=diag(W_{1}, W_{2},...,W_{T}), \text{where}
    \vspace{-2mm}
\end{equation}
\vspace{-5mm}
\begin{equation}
    W_{t}=({\mathbf{H^{l+1}_t}} A_{t_{joints}})/\sum_{i=1}^{N} {{A_{t_{joints}}}_{i}} \ 
    \vspace{-2mm}
\end{equation}

\noindent Here, $\mathbf{H^{l+1}_{i,j}}$ and $\mathbf{H_{w}^{l+1}}$ are the hyperedges and their respective weights for the next iteration. 

\noindent The process can be better understood from the algorithm \ref{algo:algo} given below. The reconstruction loss is introduced as regularization, which avoids forgetting lower-level dependencies in the embeddings $E$.
\vspace*{-1mm}
\begin{algorithm}
\caption{Forward propagation of training process }
\label{algo:algo}
\begin{algorithmic}[1]
 \For{$n \gets 1$ to \text{$total\_iteration$}}
        \If{$\text{$n$} == 1$}
            \State {$\mathbf{H^n}$} $\gets$ \text{RandomAllocationOfNodesToHyperedges} 
        
            \State {$\mathbf{H_{w}^{n}}$} $\gets$ \text{$I$(Identity Matrix)}
            
        \EndIf

        \State {$E_{enc}$} $\gets$ \text{HypEnc}($\mathbf{A},X,\mathbf{H^n},{\mathbf{H_{w}^{n}}}$)

        \State {$A_t,attn$} $\gets$ \text{HAN}($E_{enc}$), $E_f=E_{enc}+ \alpha A_t$
        \State {$rec$} $\gets$ \text{HypDec}($\text{GAP}_{\text{time}}(E_f),\mathbf{A}$),${E_c}$ $\gets$ \text{HypDec}($\text{GAP}_{\text{time}}(E_f),\mathbf{A}$)
        \State {Feed \text{$E_f$ to classification and reconstruction head}}
        \State {\text Update hypergraph for next iteration
        \State {$\mathbf{H^{n+1}}$} , {$\mathbf{H_{w}^{n+1}}$} $\gets$ \textproc{Attentive Hypergraph Generator}($attn, E_c$) }

   
 \EndFor

\Procedure{Attentive Hypergraph Generator}{$attn, E_c$}
    
    \State \textbf{Refer to section 4.2} 
    
    \State \textbf{return   } {$\mathbf{H^{n+1}}$} , {$\mathbf{H_{w}^{n+1}}$}
    
\EndProcedure
\end{algorithmic}
\end{algorithm}

\section{Loss Functions} 
\label{subsec:Loss_Functions}
\vspace*{-2mm}
Our model has encountered three losses due to vector quantization, classification, and reconstruction.

\noindent \textbf{Cross-entropy loss:} It is the measure of dissimilarity between the true distribution $y$ of labels and the predicted distribution $\hat{y}$. For a batch of $N$ skeleton sequences of $C$ various actions, the average cross-entropy loss is given by
\vspace*{-2mm}
\begin{equation}
    \mathcal{L}_{\text{CE}} = -\frac{1}{N} \sum_{n=1}^{N} \sum_{c=1}^{C} y_c^{(n)} \log(\hat{y}_c^{(n)})
    \vspace*{-2mm}
\end{equation}
\vspace*{-2mm}

\noindent \textbf{Reconstruction loss:} It is the measure of the discrepancy between the input $X$ and reconstructed output distributions $\hat{X}$, encouraging the model to accurately replicate the input skeleton sequence. The reconstruction loss is defined by
\vspace*{-2mm}
\begin{equation}
    \mathcal{L}_{\text{rec}} = \frac{1}{N} \sum_{n=1}^{N} \sum_{t=1}^{T} \sum_{j=1}^{J} \sum_{c=1}^{C} \left\| X_{t,j,c}^{(n)} - \hat{X}_{t,j,c}^{(n)} \right\|^2
    \vspace*{-2mm}
\end{equation}
\noindent where $N$, $T$, $J$, and $C$ denote the batch size, number of time frames, number of skeletal joints, and number of channels, respectively. $\| \cdot \|$ denotes the $L_2$ norm. The decoder helps regularize the generation of hypergraphs. It tries to restrict the embeddings from overfitting to high-level features and sustain some part of low-level features. As we use the similar decoder during both in-phase and out-phase hypergraph generations, two reconstruction losses $\mathcal{L}_{\text{rec1}}$ and $\mathcal{L}_{\text{rec2}}$  are introduced to the total cost objective.

\noindent \textbf{Quantization loss:} Due to the lossy quantization operation, a new loss function is added to the total cost objective, which can be defined by
\vspace{-2mm}
\begin{equation}
    \mathcal{L}_{\text{quant}} = (1/V)\sum_{i=1}^{V} \lVert Q_{he_{i}}-sg(E_{i}) \rVert_2 ^{2}.
\vspace{-2mm}
\end{equation}
``sg" denotes the non-updated stop gradient operator calculated only during the forward pass, as it has zero gradients in backpropagation. We follow the straight-through estimator approach \cite{van2017neural} for backpropagation. 

The total loss encountered during the model training is represented by 
\vspace{-2mm}
\begin{equation}
    \mathcal{L}_{\text{total}} = \mathcal{L}_{\text{CE}}+\beta_1\mathcal{L}_{\text{rec1}}+\beta_2\mathcal{L}_{\text{rec2}}+\beta_3\mathcal{L}_{\text{quant}}.
\vspace{-2mm}
\end{equation}
\noindent Here, $\beta_1$, $\beta_2$, and $\beta_3$ are the scaling factors (hyperparameters) to be optimized during model training.


\section{Experiments}
\label{sec:Experiments}
\vspace{-1mm}
\subsection{Datasets}
\label{sec:Datasets}
\vspace{-2mm}
\paragraph {NTU RGB+D 60 \cite{shahroudy2016ntu}:}It contains 56,880 sequences of 25 joints annotated for 60 action classes. The action sequences of 40 subjects are captured from three different Kinect camera angles. Cross-view (X-View) and cross-subject (X-Sub) are the protocols followed during the evaluation.
\vspace{-4.5mm}
\paragraph{NTU RGB+D 120 \cite{liu2019ntu}:} By adding 57,367 skeletons to NTU RGB+D 60, the extended dataset NTU RGB+D 120 is created by adding 60 more classes. The scenes feature 106 performers in 32 different configurations. It also uses two protocols: cross-subject (X-Sub) and cross-setup (X-Set).
\vspace{-4.5mm}
\paragraph{NW-UCLA \cite{wang2014cross}:} There are 1494 sequences in the NW-UCLA dataset that show ten distinct actions. Every sequence has twenty joint annotations for added detail. Ten people perform these actions, captured by three Kinect cameras from different perspectives.
\vspace{-2mm}
\subsection{Implementation details}
\label{sec:Implementation_details}
\vspace{-2mm}
The model is trained over 140 epochs using standard cross-entropy loss. The learning rate is initialized at 0.025 and decreased by a factor of 0.1 at the 110th and 120th epochs. We use an SGD optimizer with a Nesterov momentum value of 0.9 and a weight decay of 0.0004 during training. For NTU RGB+D and NTU RGB+D 120, a batch size of 64 is employed, with each sample resized to 64 frames, and data preprocessing follows the method described in \cite{zhang2020semantics}. Northwestern-UCLA experiments use a batch size of 16 and adhere to the data preprocessing steps outlined in \cite{chen2021channel}. The frame rate of each action sequence and the number of hyperedges are set to 64 and 5, respectively. We set the number of hyperedges, $E_h$ or K, as 5. The model architecture consists of 10 layers with 216 hidden channel dimensions for all experiments. The experiments are conducted using Python 3.10.11 based on the PyTorch 2.0.1 deep learning framework on an Ubuntu 20.04.2 machine equipped with an A100-PCIE-40GB GPU enabled with Nvidia CUDA 10.1.243 and CuDNN 8.1.0.
\vspace{-2mm}
\subsection{Ablation Analysis}
\label{sec:Ablation_Analysis}
\vspace{-2mm}
In this section, we analyze the impact of all attention techniques, hypergraph generation modules, scaling factors, channel count, and layer count of the decoder. To conduct the ablation study, we exclusively use the X-sub benchmark of the NTU RGB+D dataset.

\begin{table}[h!]
    \vspace*{-2mm}
    \captionsetup{justification=centering}
    \caption{Impact of various attention in transformer block.}\label{table:ablation_1}
    \vspace*{-2mm}
      \centering
        \begin{center}
		\vspace*{-4.5mm} 
		\scalebox{0.7}{
			\begin{tabular}{cccccc}
				\multicolumn{2}{c}{}&\textbf{Joint} &\textbf{Joint} &\textbf{Joint}  & \textbf{Accuracy}\\
                    \multicolumn{2}{c}{\textbf{Model}}&\textbf{-} &\textbf{-} &\textbf{-}  & \textbf{in}\\
                    \multicolumn{2}{c}{}&\textbf{Joint} &\textbf{Hypergraph} &\textbf{Bone}  & \textbf{(\%)}\\
				\midrule
				\multirow{5}{*}{\begin{turn}{-90} \footnotesize Base-line+\end{turn}} &\multirow{5}{*}{$\left\lbrace \begin{array}{l}
						\\
						\\
						\\
						\\
						\\
						\\
						
					\end{array}\right.$}
				&\checkmark&&&93.09    \\
				&  &&&\checkmark&93.18   \\
                    &  &&\checkmark&\checkmark&93.68 \\
                    &&\checkmark&\checkmark&&93.77   \\
				&  &\checkmark&&\checkmark&93.80   \\
				&  &\checkmark&\checkmark&\checkmark&94.15   \\[8pt]
				 \bottomrule \bottomrule
		\end{tabular}}
	\end{center}
    \vspace*{-8mm}
\end{table}

\begin{table}[h!]
\captionsetup{justification=centering}
    \caption{Impact of different units in AutoregAd-HGformer.}
    \label{table:ablation_2}
        \begin{center}
		\vspace*{-6mm}
		\scalebox{0.68}{
			\begin{tabular}{cccccc}
				\multicolumn{2}{c}{\textbf{Model}}&\textbf{out-phase} &\textbf{in-phase} &\textbf{Hypergraph}  & \textbf{Accuracy}\\
                    \multicolumn{2}{c}{}&\textbf{Hypergraph} &\textbf{Hypergraph} &\textbf{Attention}  & \textbf{in (\%)}\\
				\midrule
				\multirow{4}{*}{\begin{turn}{-90} \small Base Model +\end{turn}} &\multirow{4}{*}{$\left\lbrace \begin{array}{l}
						\\
						\\
						\\
						\\
						\\
						
					\end{array}\right.$}
				&&\checkmark&\checkmark&93.84    \\
				&  &\checkmark&&&93.65   \\
				&  &\checkmark&\checkmark&&93.88   \\
				&  &\checkmark&&\checkmark&93.79   \\
				&  &\checkmark&\checkmark&\checkmark&94.15   \\[8pt]
				\bottomrule \bottomrule
		\end{tabular}}
	\end{center}
 \vspace*{-8mm}
\end{table}
\vspace*{-3mm}
\paragraph{Impact of different attention used in transformer block:}As discussed in section \ref{sec:Hypergraph_Encoder}, we have used one self-attention and two cross-attention in the proposed AutoregAd-HGformer. Each carries different significance and weight in the output performance. As shown in Table \ref{table:ablation_1}, we calculate the model performance (accuracy in \%) of various attention in conjunction with the baseline model. The proposed model performs best when all attention is placed on a single framework.
\vspace*{-3mm}
\paragraph{Impact of different units in AutoregAd-HGformer:} Apart from the feature attention discussed in the previous paragraph, we have also adopted various hypergraph operations in the mainframe to enhance the spatiotemporal performance. These functional units are out-phase adaptive hypergraphs instead of hard-coded hypergraphs, in-phase hypergraphs, and temporal attention. We find that our in-phase and out-phase hypergraph model (93.88\%) carries a greater significance compared to the fixed hypergraph model (92.7\%) in terms of accuracy as shown in Table \ref{table:ablation_2}. The weighting of each frame according to its contribution to the final output features using temporal attention helps to neglect the insignificant frames and elevates the performance to a maximum extent of 94.15\% accuracy.
\begin{table}[h!]
    \vspace*{-4mm}
    \captionsetup{justification=centering}
    \caption{Impact of layer counts and scaling factors in  decoder.}\label{table:ablation_3}
        \begin{center}
		\vspace*{-6mm}
		\scalebox{0.85}{
			\begin{tabular}{cccccc}
				\textbf{Layer} & \multicolumn{5}{c}{\textbf{Scaling Factor}} \\ 
				\textbf{Count}&\multicolumn{1}{c}{\textbf{0.3}}  &\multicolumn{1}{c}{\textbf{0.5}} &\multicolumn{1}{c}{\textbf{0.7}} &\multicolumn{1}{c}{\textbf{0.9}} &\multicolumn{1}{c}{\textbf{1.0}} \\ \hline
				
				6&\multicolumn{1}{c}{93.62}  &\multicolumn{1}{c}{93.48} &\multicolumn{1}{c}{93.35} &\multicolumn{1}{c}{92.98} &\multicolumn{1}{c}{92.40} \\ 
				
				3&\multicolumn{1}{c}{93.69}  &\multicolumn{1}{c}{93.84} &\multicolumn{1}{c}{93.97} &\multicolumn{1}{c}{94.15} &\multicolumn{1}{c}{93.92} \\ \hline
				\bottomrule
		\end{tabular}}
	\end{center}
\vspace*{-6mm}
\end{table}

\begin{table}[h!]
    \vspace*{-5mm}
    \captionsetup{justification=centering}
    \caption{Effect of input dimension fed to hypergraph generator.}\label{table:ablation_4}
        \begin{center}
		\vspace*{-6mm}
		\scalebox{0.85}{
		\begin{tabular}{ccccc}
			\textbf{Channel Count}&\textbf{4}  &\textbf{6} &\textbf{8}&\textbf{12}\\ 
                \hline	
			Accuracy (\%)&93.68  &93.90 &94.15 &93.86 \\ \hline\hline
		\end{tabular}}
	\end{center}
    \vspace*{-6mm}
\end{table}

\vspace*{-6mm}
\begin{table}
    \vspace*{-4mm}
    \captionsetup{justification=centering}
    \caption{Impact of the scaling factor as shown in Fig. \ref{fig:WACV_Model}}.
    \label{table:ablation_5}
        \begin{center}
		\vspace*{-6mm}
		\scalebox{0.9}{
		\begin{tabular}{cccccc}
			\textbf{ $\alpha$ }&\textbf{0.1}  &\textbf{0.2} &\textbf{0.3} &\textbf{0.4}&\textbf{0.6}\\ 
                \hline	
			Accuracy (\%)&93.64  &94.15 &93.92 &93.76 &93.67 \\ \hline \hline
		\end{tabular}}
	\end{center}
 \vspace*{-4mm}
\end{table} 

\begin{table*}[h!]
    \vspace*{-3mm}
	\begin{center}
		\captionsetup{justification=centering}
		\caption{Quantitative comparison of AutoregAd-HGformer on NTU RGB+D 60, NTU RGB+D 120, and NW-UCLA datasets.}
		\vspace*{-2mm}
		\tabcolsep=0.11cm
		\scalebox{0.8}{
			\begin{tabular}{ccccccc}
				\multirow{2}{*}{\textbf{Architecture}} & \multirow{2}{*}{\textbf{Method}} &  \multicolumn{2}{c}{\textbf{NTU-60}} & \multicolumn{2}{c}{\textbf{NTU-120}} & \multirow{2}{*}{\textbf{NW-UCLA}}   \\ \cmidrule(lr){3-4}\cmidrule(lr){5-6}
				&  &\multicolumn{1}{c}{X-Sub (\%)} &\multicolumn{1}{c}{X-View (\%)} &\multicolumn{1}{c}{X-Sub (\%)} &\multicolumn{1}{c}{X-set (\%)}& \\ \hline
				\multirow{8}{*}{Graph Convolution}&ST-GCN \cite{yan2018spatial}           &\multicolumn{1}{c}{81.5}&\multicolumn{1}{c}{88.3}&\multicolumn{1}{c}{70.7}&\multicolumn{1}{c}{73.2} &-   \\
				&2S-AGCN \cite{shi2019two}  &\multicolumn{1}{c}{88.5}&\multicolumn{1}{c}{95.1}&\multicolumn{1}{c}{82.5}&\multicolumn{1}{c}{84.2} & -  \\
				&Shift-GCN \cite{cheng2020skeleton}       &\multicolumn{1}{c}{90.7}&\multicolumn{1}{c}{96.5}&\multicolumn{1}{c}{85.9}&\multicolumn{1}{c}{87.6} &94.6   \\
				&SGN \cite{zhang2020semantics}    &\multicolumn{1}{c}{89.0}&\multicolumn{1}{c}{94.5}&\multicolumn{1}{c}{79.2}&\multicolumn{1}{c}{81.5} &92.5   \\
				&CTR-GCN \cite{chen2021channel}         &\multicolumn{1}{c}{92.4}&\multicolumn{1}{c}{96.8}&\multicolumn{1}{c}{88.9}&\multicolumn{1}{c}{90.6} &96.5   \\
				&Info-GCN \cite{chi2022infogcn}             
				& \multicolumn{1}{c}{93.0}&\multicolumn{1}{c}{97.1}&\multicolumn{1}{c}{89.8}&\multicolumn{1}{c}{91.2}&97.0\\
				& HLP-GCN \cite{wei2023accommodating}       &\multicolumn{1}{c}{92.7}&\multicolumn{1}{c}{96.9}&\multicolumn{1}{c}{89.0}&\multicolumn{1}{c}{90.8} & 96.8   \\
				&HD-GCN \cite{lee2023hierarchically}             
				& \multicolumn{1}{c}{93.4}&\multicolumn{1}{c}{97.2}&\multicolumn{1}{c}{90.1}&\multicolumn{1}{c}{91.6}&97.2\\ 
				
				\hdashline      
				\multirow{6}{*}{Graph Transformer}&ST-TR \cite{plizzari2021spatial}  &\multicolumn{1}{c}{89.9}&\multicolumn{1}{c}{96.1}&\multicolumn{1}{c}{81.9}&\multicolumn{1}{c}{84.1} &-   \\
				&STST \cite{zhang2021stst}       &\multicolumn{1}{c}{91.9}&\multicolumn{1}{c}{96.8}&\multicolumn{1}{c}{-}&\multicolumn{1}{c}{-}  & -  \\
				&IIP-Transformer \cite{wang2023iip}    &\multicolumn{1}{c}{92.3}&\multicolumn{1}{c}{96.4}&\multicolumn{1}{c}{88.4}&\multicolumn{1}{c}{89.7}  &-   \\
				&FG-STFormer \cite{gao2022focal}         &\multicolumn{1}{c}{92.6}&\multicolumn{1}{c}{96.7}&\multicolumn{1}{c}{89.0}&\multicolumn{1}{c}{90.6} &97.0   \\
				&IGFormer \cite{pang2022igformer}       &\multicolumn{1}{c}{93.4}&\multicolumn{1}{c}{96.5}&\multicolumn{1}{c}{85.4}&\multicolumn{1}{c}{86.5} &-   \\
				&MAMP \cite{mao2023masked}             
				&\multicolumn{1}{c}{93.1}&\multicolumn{1}{c}{97.5}&\multicolumn{1}{c}{90.0}&\multicolumn{1}{c}{91.3}&-\\
				
				\hdashline
				\multirow{6}{*}{Hypergraph Convolution}&Hyper-GNN \cite{hao2021hypergraph}       &\multicolumn{1}{c}{89.5}&\multicolumn{1}{c}{95.7}&\multicolumn{1}{c}{-}&\multicolumn{1}{c}{-}  &-   \\
				&DH-GCN \cite{cheng2020decoupling}    &\multicolumn{1}{c}{90.7}&\multicolumn{1}{c}{96.0}&\multicolumn{1}{c}{86.0}&\multicolumn{1}{c}{87.9} &-   \\
				&Selective-HCN \cite{zhu2022selective}         &\multicolumn{1}{c}{90.8}&\multicolumn{1}{c}{96.6}&\multicolumn{1}{c}{-}&\multicolumn{1}{c}{-}  & -  \\
				&DST-HCN \cite{wang2023dynamic}       &\multicolumn{1}{c}{92.3}&\multicolumn{1}{c}{96.8}&\multicolumn{1}{c}{88.8}&\multicolumn{1}{c}{90.7} &96.6   \\				
				\hdashline 
				\multirow{2}{*}{Hypergraph Transformer}&Hyperformer \cite{zhou2022hypergraph}       &\multicolumn{1}{c}{92.9}&\multicolumn{1}{c}{96.5}&\multicolumn{1}{c}{89.9}&\multicolumn{1}{c}{91.3} &\multicolumn{1}{c}{96.9}   \\
				&\textbf{AutoregAd-HGformer (ours) }          
				& \multicolumn{1}{c}{\textbf{94.15}}&\multicolumn{1}{c}{\textbf{97.83}}&\multicolumn{1}{c}{\textbf{91.02}}&\multicolumn{1}{c}{\textbf{92.42}}&\multicolumn{1}{c}{\textbf{97.98}}\\
				\bottomrule \bottomrule
		\end{tabular}}
		\label{tab:Ad-HGformer_quant}
	\end{center}
    \vspace*{-9mm} 
\end{table*}

\paragraph{Impact of layer counts and scaling factors in decoder:} We apply hybrid learning, combining supervised and unsupervised techniques, to enhance the performance of the proposed AutoregAd-HGformer. A decoder is used on the final classification feature embedding to apply unsupervised learning. As shown in Table \ref{table:ablation_3}, we evaluate the optimal layer count and scaling factor for reconstruction error related to the decoder unit. We observe that the three-layered decoder with a scaling factor of 0.9 yields the highest accuracy of 94.15\%. We also observe that a smaller scaling factor should be chosen for a larger-layered decoder for optimal performance and vice versa.
\vspace*{-3mm}
\paragraph{Effect of input dimension $d$ fed to hypergraph generator and scaling factor $\alpha$:} The k-means clustering would not perform best for the embedding either with larger dimensions or with smaller dimensions due to ``Curse of Dimensionality". As shown in Table \ref{table:ablation_4}, we also observe that AutoregAd-HGformer yields the highest performance in terms of accuracy (94.15\%) for 8-channel embedding compared to 4, 6, and 12. Similarly, as shown in Table \ref{table:ablation_5}, the proposed model yields the highest performance for the scaling factor $\alpha=0.2$ compared to 0.1, 0.3, 0.4, and 0.6.

\vspace*{-1mm}

\subsection{Quantitative Comparison}
\label{sec:Quantitative_Comparison}
\vspace*{-2mm} 
Table \ref{tab:Ad-HGformer_quant} provides a comprehensive comparison of our AutoregAd-HGformer model with state-of-the-art methods across four different architectures: Graph Convolution, Graph Transformer, Hypergraph Convolution, and Hypergraph Transformer. The comparison is based on the top-1 accuracy (\%) on three datasets: NTU RGB+D 60, NTU RGB+D 120, and NW-UCLA.
\begin{enumerate}
	\item[\ding{172}] Graph Convolution Architecture: As shown in Table \ref{tab:Ad-HGformer_quant}, the recent HD-GCN method is very competitive compared to the proposed transformer-based AutoregAd-HGformer. However, AutoregAd-HGformer achieves a top-1 accuracy of 94.15\% and 97.83\% on two settings of NTU RGB+D 60, 91.02\% and 92.42\% on two settings of NTU RGB+D 120, and 97.98\% on NW-UCLA datasets, outperforming all other methods based on this architecture.
	\item[\ding{173}] Graph Transformer Architecture: As the proposed model adopts the hypergraph transformer technique in its mainframe, we compare its performance to SOTA graph transformer architecture. AutoregAd-HGformer continues to perform exceptionally well in this architecture on each dataset. 
	\item[\ding{174}] Hypergraph Convolution Architecture: The conceptualization of the proposed model is centered around hypergraph by grouping the contextually correlated graph nodes in the graph. AutoregAd-HGformer also maintains its superiority with accuracy on the NTU RGB+D 60, NTU RGB+D 120, and NW-UCLA datasets in these settings.
	\item[\ding{175}] Hypergraph Transformer Architecture: As little attention has been paid to hypergraph transformers so far, we found only one paper implemented called Hyperformer. The performance of AutoregAd-HGformer also surpasses the state-of-the-art hypergraph transformer-based architecture by a margin of 0.7\% in terms of top-1 accuracy.
\end{enumerate}
Overall, our AutoregAd-HGformer model consistently outperforms state-of-the-art methods across all architectures and datasets, demonstrating its superior performance and effectiveness for action recognition tasks.

\begin{table}[h!]
\captionsetup{justification=centering}
\caption{Model-agnostic performance of hypergraph decoder.}
\label{table:model-agnostic}
\vspace*{-6mm} 
\begin{center}
\scalebox{0.65}{
\begin{tabular}{cccccc}
\textbf{Dataset} &\multicolumn{2}{c}{NTU-60}&\multicolumn{2}{c}{NTU-120}&NW-UCLA                        \\ \hline
\textbf{Setting}    &\textbf{X-Sub}&\textbf{X-View}&\textbf{X-Sub}&\textbf{X-View}& \\ \hline 
DST-HCN\cite{wang2023dynamic}  & 90.7 & 96.0 & 86.0 & 87.9  & - \\
DST-HCN*  & 90.7 & 95.9 & 85.9 & 87.9 & 94.9 \\
\textbf{DST-HCN*+Ad HypDec} & \textbf{91.3} & \textbf{96.4} & \textbf{86.5}  & \textbf{88.3} & \textbf{95.1}\\ \hdashline 
Selective-HCN \cite{zhu2022selective}  & 90.8       & 96.6  & - & -  & - \\
Selective-HCN*  & 90.7   & 96.6 & 86.3   & 88.1 & 95.0    \\
\textbf{Selective-HCN*+Ad HypDec} & \textbf{91.4}& \textbf{97.0} & \textbf{86.7} & \textbf{88.3} & \textbf{95.3}\\ \hdashline 
Hyperformer\cite{zhou2022hypergraph}  & 92.9  & 96.5  & 89.9  & 91.3  & 96.9  \\
Hyperformer* & 92.9  & 96.4 & 89.9 & 91.3 & 96.8 \\
\textbf{Hyperformer*+Ad HypDec} & \textbf{93.65} & \textbf{97.25}  & \textbf{90.49}  & \textbf{91.91} & \textbf{97.43}\\ \hdashline 
3Mformer\cite{wang20233mformer}  & 94.8  & 98.7  & 92.0  & 93.8  & 97.8  \\
3Mformer* & 94.8  & 98.6 & 91.9 & 93.8 & 97.8 \\
\textbf{3Mformer*+Ad HypDec} & \textbf{95.2} & \textbf{98.9}  & \textbf{92.2}  & \textbf{94.1} & \textbf{98.1}\\ \bottomrule \bottomrule
\end{tabular}}
\end{center}
 \vspace{-4mm}
\end{table}

\vspace*{-1mm} 
\subsection{Model-agnostic Performance Comparison }
\label{sec:Qualitative_Comparison}
\vspace*{-1mm} 
Instead of taking only the Hyperformer \cite{zhou2022hypergraph} as the base encoder, we have also considered DST-HCN \cite{wang2023dynamic}, Selective-HCN \cite{zhu2022selective}, and 3Mformer \cite{wang20233mformer} as the hypergraph encoder and observed the enhancement in model-agnostic performance. As clearly observed in Table \ref{table:model-agnostic}, all the listed hypergraph encoders with the proposed adaptive hypergraph decoder surpass the solo individual performances.

\vspace*{-1mm}
\begin{figure}[ht]
    \vspace*{-2mm}
    \centering
    \captionsetup{justification=centering}
    \includegraphics[width=0.97\linewidth]{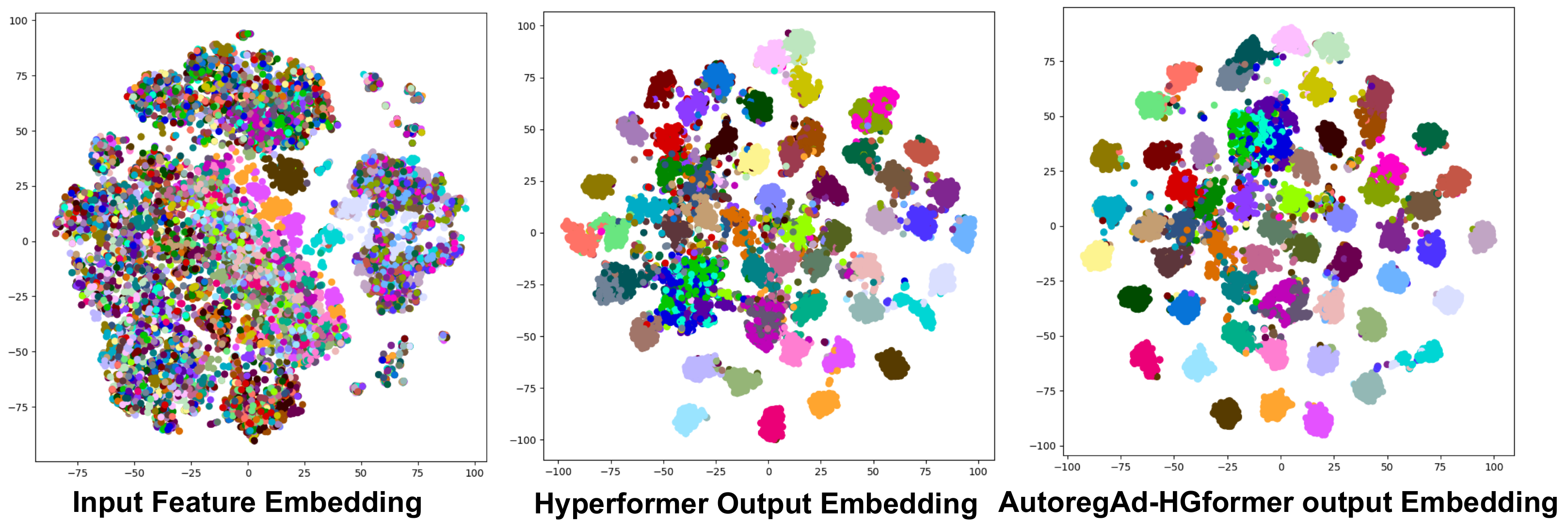}
    \vspace*{-2.5mm}
    \caption{ t-SNE \cite{van2008visualizing} of input features (left) and model output feature embeddings.}
    \label{fig:tsne_adhgformer}
    \vspace*{-6mm}
\end{figure}
\subsection{Qualitative Comparison}
\label{sec:Qualitative_Comparison}
\vspace*{-1mm} 
In Fig. \ref{fig:tsne_adhgformer}, we present the t-SNE distribution plot of output embeddings of Hyperformer (middle) and AutoregAd-HGformer (right) trained on the NTU RGB+D 60 dataset. The more organized clustering of skeleton-based action data points in Ad-HFformer highlights its superiority over other hypergraph transformer-based  Hyperformer architectures. This figure also indicates that the proposed method can handle skeletal action sequences more effectively by maximizing the intra-class similarity and inter-class separability features, i.e., skeleton-based action sequences from the same class clustering together while those from different classes spread apart.
\vspace*{-1mm}

\subsection{Model complexity}
\label{sec:Model_complexity}
The tradeoff between computational complexity and model performance (accuracy in \%) is illustrated in Table \ref{table:complexity}. Our model performs better with fewer parameters than CTR-GCN, Info-GCN, HD-GCN, and DST-CN. Although the proposed AutoregAd-HGformer architecture needs more parameters than ST-GCN and Shift-GCN, it outperforms the above two by a large margin in terms of flops and accuracies. The above ablation is analyzed on the NTU RGB+D 60 dataset.

\begin{table}[h!]
	\begin{center}
		\tabcolsep=0.13cm
		\vspace*{-2mm}
		\centering
		\captionsetup{justification=centering}
		\caption{Performance vs Model Complexity.}
		\vspace*{-3mm}
		\scalebox{0.7}{
			\begin{tabular}{ccccc}
				Models & Publication & Params (M) & Flops (G) & X-Sub/X-View (\%) \\ \hline 
				ST-GCN \cite{yan2018spatial}  & AAAI-2018 & 3.08 & 16.32 & 81.5/88.3 \\
                    Shift-GCN \cite{cheng2020skeleton}  & CVPR-2020 & 2.76 & 10.01 & 90.7/96.5 \\
				CTR-GCN \cite{chen2021channel}  & ICCV-2021  & 5.84  & 7.88 & 92.4/96.8 \\
				Info-GCN \cite{chi2022infogcn}  & ICCV-2022  & 6.28 & 6.72 & 92.7/96.9 \\
				HD-GCN \cite{lee2023hierarchically}  & ICCV-2023  & 6.72 & 6.40 & 93.0/97.0 \\
				DST-HCN \cite{wang2023dynamic}  & ICME-2023  & 3.50  & 2.93 & 92.3/96.8 \\
				Hyperformer \cite{cheng2020skeleton}  & arXiv-2023  & 2.60 & 14.8 & 92.9/96.5 \\
				\textbf{AutoregAd-HGformer }  & \textbf{Proposed}  & \textbf{3.20} & \textbf{15.4} & \textbf{94.15}/\textbf{97.83} \\
				\bottomrule
				\bottomrule
		\end{tabular}}
		\label{table:complexity}
	\end{center}
	\vspace*{-7mm}
\end{table}

\section{Misclassification}
\label{sec:Misclassification}
We observe the semantics of some action sequences in NTU RGB+D 60 datasets are very similar and often misclassified by the present state-of-the-art models. For example: (a) ``writing[A12]" as ``type on a keyboard[A30]" and vice-versa, (b) ``reading[A11]" as ``writing[A12]" and vice-versa, (c) ``reading[A11]" as ``play with the phone/tablet[A29]" and vice-versa, and (d) ``writing[A12]" as ``play with the phone/tablet[A29]" and vice-versa. We also recognize that using the proposed adaptive hyperedge decoder helps competently reduce the misclassification of the above action sets. Fig. \ref{fig:bargraph1} validates the significance of the adaptive hyperedge decoder in the proposed AutoregAd-HGformer.
\begin{figure}[h]
    \vspace*{-2mm}
    \centering
    \captionsetup{justification=centering}
    \includegraphics[width=1.0\linewidth]{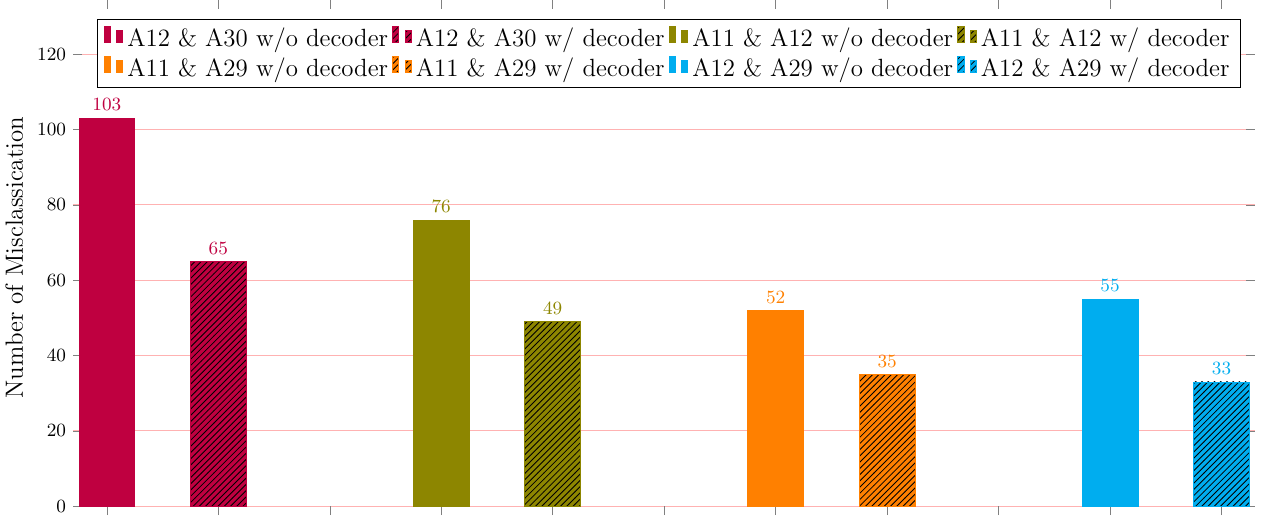}
    \vspace*{-5mm}
    \caption{Misclassification between various ambiguous actions (Axx \& Axx) of NTU RGB+D 60 dataset before and after implementing adaptive decoder. A11:reading, A12:writing, A29:play with the phone/tablet, A30:type on a keyboard.}
    \label{fig:bargraph1}
    \vspace*{-6mm}
\end{figure}


\section{Conclusion}
\label{sec:Conclusion}

We propose a novel adaptive hypergraph transformer framework called AutoregAd-HGformer that aggregates multiscale graphs and higher-order contextual semantics with long-range motion features to recognize skeleton sequences for diverse actions. We apply attention mechanisms such as joint-joint self-attention, joint-hyperedge cross-attention, and joint-bone cross-attention to distill frame-level and temporal attention to derive motion-level deep action-dependent features. We also employ channel attention to capture heterogeneous contextual information. The post-extraction decoder helps execute hybrid learning (supervised and self-supervised) and iterative hyperedge clustering to make the model more robust. The novel vector quantized in-phase and model-agnostic out-phase hypergraph generation helps the model to aggregate more robust features for accurate hyperedge calculation. Extensive experiments on multiple datasets show the effectiveness of AutoregAd-HGformer. By incorporating post-extraction adaptive hypergraphs with joint, hyperedge, bone, node, and channel-level self- or cross-attention, AutoregAd-HGformer outperforms the other state-of-the-art architectures on the NTU RGB+D 60, NTU RGB+D 120, and NW-UCLA datasets. The hyperedge-hyperedge self-attention is left for future consideration. 

{\small
\bibliographystyle{ieee_fullname}
\bibliography{egbib}

\begin{thebibliography}{10}\itemsep=-1pt

\bibitem{aksan2021spatio}
Emre Aksan, Manuel Kaufmann, Peng Cao, and Otmar Hilliges.
\newblock A spatio-temporal transformer for 3d human motion prediction.
\newblock In {\em 2021 International Conference on 3D Vision (3DV)}, pages
  565--574. IEEE, 2021.

\bibitem{bengio2013estimating}
Yoshua Bengio, Nicholas L{\'e}onard, and Aaron Courville.
\newblock Estimating or propagating gradients through stochastic neurons for
  conditional computation.
\newblock {\em arXiv preprint arXiv:1308.3432}, 2013.

\bibitem{chandra2023escaping}
Noirrit~Kiran Chandra, Antonio Canale, and David~B Dunson.
\newblock Escaping the curse of dimensionality in bayesian model-based
  clustering.
\newblock {\em Journal of machine learning research}, 24(144):1--42, 2023.

\bibitem{chen2022activating}
Xiangyu Chen, Xintao Wang, Jiantao Zhou, and Chao Dong.
\newblock Activating more pixels in image super-resolution transformer. arxiv
  2022.
\newblock {\em arXiv preprint arXiv:2205.04437}, 1, 2022.

\bibitem{chen2021channel}
Yuxin Chen, Ziqi Zhang, Chunfeng Yuan, Bing Li, Ying Deng, and Weiming Hu.
\newblock Channel-wise topology refinement graph convolution for skeleton-based
  action recognition.
\newblock In {\em Proceedings of the IEEE/CVF international conference on
  computer vision}, pages 13359--13368, 2021.

\bibitem{cheng2020decoupling}
Ke Cheng, Yifan Zhang, Congqi Cao, Lei Shi, Jian Cheng, and Hanqing Lu.
\newblock Decoupling gcn with dropgraph module for skeleton-based action
  recognition.
\newblock In {\em Computer Vision--ECCV 2020: 16th European Conference,
  Glasgow, UK, August 23--28, 2020, Proceedings, Part XXIV 16}, pages 536--553.
  Springer, 2020.

\bibitem{cheng2020skeleton}
Ke Cheng, Yifan Zhang, Xiangyu He, Weihan Chen, Jian Cheng, and Hanqing Lu.
\newblock Skeleton-based action recognition with shift graph convolutional
  network.
\newblock In {\em Proceedings of the IEEE/CVF conference on computer vision and
  pattern recognition}, pages 183--192, 2020.

\bibitem{chi2022infogcn}
Hyung-gun Chi, Myoung~Hoon Ha, Seunggeun Chi, Sang~Wan Lee, Qixing Huang, and
  Karthik Ramani.
\newblock Infogcn: Representation learning for human skeleton-based action
  recognition.
\newblock In {\em Proceedings of the IEEE/CVF Conference on Computer Vision and
  Pattern Recognition}, pages 20186--20196, 2022.

\bibitem{cho2020self}
Sangwoo Cho, Muhammad Maqbool, Fei Liu, and Hassan Foroosh.
\newblock Self-attention network for skeleton-based human action recognition.
\newblock In {\em Proceedings of the IEEE/CVF Winter Conference on Applications
  of Computer Vision}, pages 635--644, 2020.

\bibitem{gao2022focal}
Zhimin Gao, Peitao Wang, Pei Lv, Xiaoheng Jiang, Qidong Liu, Pichao Wang,
  Mingliang Xu, and Wanqing Li.
\newblock Focal and global spatial-temporal transformer for skeleton-based
  action recognition.
\newblock In {\em Proceedings of the Asian Conference on Computer Vision},
  pages 382--398, 2022.

\bibitem{hao2021hypergraph}
Xiaoke Hao, Jie Li, Yingchun Guo, Tao Jiang, and Ming Yu.
\newblock Hypergraph neural network for skeleton-based action recognition.
\newblock {\em IEEE Transactions on Image Processing}, 30:2263--2275, 2021.

\bibitem{hu2019squeezeandexcitationnetworks}
Jie Hu, Li Shen, Samuel Albanie, Gang Sun, and Enhua Wu.
\newblock Squeeze-and-excitation networks, 2019.

\bibitem{ke2017new}
Qiuhong Ke, Mohammed Bennamoun, Senjian An, Ferdous Sohel, and Farid Boussaid.
\newblock A new representation of skeleton sequences for 3d action recognition.
\newblock In {\em Proceedings of the IEEE conference on computer vision and
  pattern recognition}, pages 3288--3297, 2017.

\bibitem{lee2023hierarchically}
Jungho Lee, Minhyeok Lee, Dogyoon Lee, and Sangyoun Lee.
\newblock Hierarchically decomposed graph convolutional networks for
  skeleton-based action recognition.
\newblock In {\em Proceedings of the IEEE/CVF International Conference on
  Computer Vision}, pages 10444--10453, 2023.

\bibitem{li2017skeleton}
Chao Li, Qiaoyong Zhong, Di Xie, and Shiliang Pu.
\newblock Skeleton-based action recognition with convolutional neural networks.
\newblock In {\em 2017 IEEE international conference on multimedia \& expo
  workshops (ICMEW)}, pages 597--600. IEEE, 2017.

\bibitem{liu2019ntu}
Jun Liu, Amir Shahroudy, Mauricio Perez, Gang Wang, Ling-Yu Duan, and Alex~C
  Kot.
\newblock Ntu rgb+ d 120: A large-scale benchmark for 3d human activity
  understanding.
\newblock {\em IEEE transactions on pattern analysis and machine intelligence},
  42(10):2684--2701, 2019.

\bibitem{liu2016spatio}
Jun Liu, Amir Shahroudy, Dong Xu, and Gang Wang.
\newblock Spatio-temporal lstm with trust gates for 3d human action
  recognition.
\newblock In {\em Computer Vision--ECCV 2016: 14th European Conference,
  Amsterdam, The Netherlands, October 11-14, 2016, Proceedings, Part III 14},
  pages 816--833. Springer, 2016.

\bibitem{long2023step}
Nguyen Huu~Bao Long.
\newblock Step catformer: Spatial-temporal effective body-part cross attention
  transformer for skeleton-based action recognition.
\newblock {\em arXiv preprint arXiv:2312.03288}, 2023.

\bibitem{mao2023masked}
Yunyao Mao, Jiajun Deng, Wengang Zhou, Yao Fang, Wanli Ouyang, and Houqiang Li.
\newblock Masked motion predictors are strong 3d action representation
  learners.
\newblock In {\em Proceedings of the IEEE/CVF International Conference on
  Computer Vision}, pages 10181--10191, 2023.

\bibitem{pang2022igformer}
Yunsheng Pang, Qiuhong Ke, Hossein Rahmani, James Bailey, and Jun Liu.
\newblock Igformer: Interaction graph transformer for skeleton-based human
  interaction recognition.
\newblock In {\em European Conference on Computer Vision}, pages 605--622.
  Springer, 2022.

\bibitem{plizzari2021spatial}
Chiara Plizzari, Marco Cannici, and Matteo Matteucci.
\newblock Spatial temporal transformer network for skeleton-based action
  recognition.
\newblock In {\em Pattern Recognition. ICPR International Workshops and
  Challenges: Virtual Event, January 10--15, 2021, Proceedings, Part III},
  pages 694--701. Springer, 2021.

\bibitem{shahroudy2016ntu}
Amir Shahroudy, Jun Liu, Tian-Tsong Ng, and Gang Wang.
\newblock Ntu rgb+ d: A large scale dataset for 3d human activity analysis.
\newblock In {\em Proceedings of the IEEE conference on computer vision and
  pattern recognition}, pages 1010--1019, 2016.

\bibitem{shi2019two}
Lei Shi, Yifan Zhang, Jian Cheng, and Hanqing Lu.
\newblock Two-stream adaptive graph convolutional networks for skeleton-based
  action recognition.
\newblock In {\em Proceedings of the IEEE/CVF conference on computer vision and
  pattern recognition}, pages 12026--12035, 2019.

\bibitem{shi2020decoupled}
Lei Shi, Yifan Zhang, Jian Cheng, and Hanqing Lu.
\newblock Decoupled spatial-temporal attention network for skeleton-based
  action-gesture recognition.
\newblock In {\em Proceedings of the Asian Conference on Computer Vision},
  2020.

\bibitem{song2022constructing}
Yi-Fan Song, Zhang Zhang, Caifeng Shan, and Liang Wang.
\newblock Constructing stronger and faster baselines for skeleton-based action
  recognition.
\newblock {\em IEEE transactions on pattern analysis and machine intelligence},
  45(2):1474--1488, 2022.

\bibitem{soo2017interpretable}
Tae Soo~Kim and Austin Reiter.
\newblock Interpretable 3d human action analysis with temporal convolutional
  networks.
\newblock In {\em Proceedings of the IEEE conference on computer vision and
  pattern recognition workshops}, pages 20--28, 2017.

\bibitem{tran2018closer}
Du Tran, Heng Wang, Lorenzo Torresani, Jamie Ray, Yann LeCun, and Manohar
  Paluri.
\newblock A closer look at spatiotemporal convolutions for action recognition.
\newblock In {\em Proceedings of the IEEE conference on Computer Vision and
  Pattern Recognition}, pages 6450--6459, 2018.

\bibitem{van2017neural}
Aaron Van Den~Oord, Oriol Vinyals, et~al.
\newblock Neural discrete representation learning.
\newblock {\em Advances in neural information processing systems}, 30, 2017.

\bibitem{van2008visualizing}
Laurens Van~der Maaten and Geoffrey Hinton.
\newblock Visualizing data using t-sne.
\newblock {\em Journal of machine learning research}, 9(11), 2008.

\bibitem{wang2014cross}
Jiang Wang, Xiaohan Nie, Yin Xia, Ying Wu, and Song-Chun Zhu.
\newblock Cross-view action modeling, learning and recognition.
\newblock In {\em Proceedings of the IEEE conference on computer vision and
  pattern recognition}, pages 2649--2656, 2014.

\bibitem{wang2019comparative}
Lei Wang, Du~Q Huynh, and Piotr Koniusz.
\newblock A comparative review of recent kinect-based action recognition
  algorithms.
\newblock {\em IEEE Transactions on Image Processing}, 29:15--28, 2019.

\bibitem{wang20233mformer}
Lei Wang and Piotr Koniusz.
\newblock 3mformer: Multi-order multi-mode transformer for skeletal action
  recognition.
\newblock In {\em Proceedings of the IEEE/CVF Conference on Computer Vision and
  Pattern Recognition}, pages 5620--5631, 2023.

\bibitem{wang2015action}
Limin Wang, Yu Qiao, and Xiaoou Tang.
\newblock Action recognition with trajectory-pooled deep-convolutional
  descriptors.
\newblock In {\em Proceedings of the IEEE conference on computer vision and
  pattern recognition}, pages 4305--4314, 2015.

\bibitem{wang2023iip}
Qingtian Wang, Shuze Shi, Jiabin He, Jianlin Peng, Tingxi Liu, and Renliang
  Weng.
\newblock Iip-transformer: Intra-inter-part transformer for skeleton-based
  action recognition.
\newblock In {\em 2023 IEEE International Conference on Big Data (BigData)},
  pages 936--945. IEEE, 2023.

\bibitem{10220028}
S. Wang, Y. Zhang, H. Qi, M. Zhao, and Y. Jiang.
\newblock Dynamic spatial-temporal hypergraph convolutional network for
  skeleton-based action recognition.
\newblock In {\em 2023 IEEE International Conference on Multimedia and Expo
  (ICME)}, pages 2147--2152, Los Alamitos, CA, USA, jul 2023. IEEE Computer
  Society.

\bibitem{wang2023dynamic}
Shengqin Wang, Yongji Zhang, Hong Qi, Minghao Zhao, and Yu Jiang.
\newblock Dynamic spatial-temporal hypergraph convolutional network for
  skeleton-based action recognition.
\newblock {\em arXiv preprint arXiv:2302.08689}, 2023.

\bibitem{wei2023accommodating}
Chao Wei and Zhidong Deng.
\newblock Accommodating self-attentional heterophily topology into high-and
  low-pass graph convolutional network for skeleton-based action recognition.
\newblock In {\em 2023 International Joint Conference on Neural Networks
  (IJCNN)}, pages 01--08. IEEE, 2023.

\bibitem{wei2021dynamic}
Jinfeng Wei, Yunxin Wang, Mengli Guo, Pei Lv, Xiaoshan Yang, and Mingliang Xu.
\newblock Dynamic hypergraph convolutional networks for skeleton-based action
  recognition.
\newblock {\em arXiv preprint arXiv:2112.10570}, 2021.

\bibitem{wu2023skeletonmae}
Wenhan Wu, Yilei Hua, Ce Zheng, Shiqian Wu, Chen Chen, and Aidong Lu.
\newblock Skeletonmae: Spatial-temporal masked autoencoders for self-supervised
  skeleton action recognition.
\newblock In {\em 2023 IEEE International Conference on Multimedia and Expo
  Workshops (ICMEW)}, pages 224--229. IEEE, 2023.

\bibitem{xin2023transformer}
Wentian Xin, Ruyi Liu, Yi Liu, Yu Chen, Wenxin Yu, and Qiguang Miao.
\newblock Transformer for skeleton-based action recognition: A review of recent
  advances.
\newblock {\em Neurocomputing}, 2023.

\bibitem{yan2018spatial}
Sijie Yan, Yuanjun Xiong, and Dahua Lin.
\newblock Spatial temporal graph convolutional networks for skeleton-based
  action recognition.
\newblock In {\em Proceedings of the AAAI conference on artificial
  intelligence}, volume~32, 2018.

\bibitem{zhang2020semantics}
Pengfei Zhang, Cuiling Lan, Wenjun Zeng, Junliang Xing, Jianru Xue, and Nanning
  Zheng.
\newblock Semantics-guided neural networks for efficient skeleton-based human
  action recognition.
\newblock In {\em proceedings of the IEEE/CVF conference on computer vision and
  pattern recognition}, pages 1112--1121, 2020.

\bibitem{zhang2017geometric}
Songyang Zhang, Xiaoming Liu, and Jun Xiao.
\newblock On geometric features for skeleton-based action recognition using
  multilayer lstm networks.
\newblock In {\em 2017 IEEE winter conference on applications of computer
  vision (WACV)}, pages 148--157. IEEE, 2017.

\bibitem{zhang2021stst}
Yuhan Zhang, Bo Wu, Wen Li, Lixin Duan, and Chuang Gan.
\newblock Stst: Spatial-temporal specialized transformer for skeleton-based
  action recognition.
\newblock In {\em Proceedings of the 29th ACM International Conference on
  Multimedia}, pages 3229--3237, 2021.

\bibitem{zhao2017temporal}
Yue Zhao, Yuanjun Xiong, Limin Wang, Zhirong Wu, Xiaoou Tang, and Dahua Lin.
\newblock Temporal action detection with structured segment networks.
\newblock In {\em Proceedings of the IEEE international conference on computer
  vision}, pages 2914--2923, 2017.

\bibitem{zhou2022hypergraph}
Yuxuan Zhou, Zhi-Qi Cheng, Chao Li, Yanwen Fang, Yifeng Geng, Xuansong Xie, and
  Margret Keuper.
\newblock Hypergraph transformer for skeleton-based action recognition.
\newblock {\em arXiv preprint arXiv:2211.09590}, 2022.

\bibitem{zhu2016co}
Wentao Zhu, Cuiling Lan, Junliang Xing, Wenjun Zeng, Yanghao Li, Li Shen, and
  Xiaohui Xie.
\newblock Co-occurrence feature learning for skeleton based action recognition
  using regularized deep lstm networks.
\newblock In {\em Proceedings of the AAAI conference on artificial
  intelligence}, volume~30, 2016.

\bibitem{zhu2022selective}
Yiran Zhu, Guangji Huang, Xing Xu, Yanli Ji, and Fumin Shen.
\newblock Selective hypergraph convolutional networks for skeleton-based action
  recognition.
\newblock In {\em Proceedings of the 2022 International Conference on
  Multimedia Retrieval}, pages 518--526, 2022.

\end{thebibliography}
}

\newpage

\FloatBarrier
\begin{table*}[h!]
	\begin{center}
		\tabcolsep=0.13cm
		\vspace*{-2mm}
		\centering
		\captionsetup{justification=centering}
		\vspace*{-3mm}
		\scalebox{0.7}{
			\begin{tabular}{c}
				\Huge Autoregressive Adaptive Hypergraph Transformer for \\ \Huge Skeleton-based Activity Recognition\\ \\ \LARGE -:Supplemetary Material:-
		\end{tabular}}
	\end{center}
	\vspace*{-7mm}
\end{table*}
\FloatBarrier

This supplementary material provides additional information about hypergraph convolution, temporal attention in STA-HT block, hypergraph quantizer, comprehensive ablation study, and additional experimental results in sec. \ref{sec:Hypergraph_Convolution}, \ref{sec:Temporal_attention}, \ref{sec:Hypergraph_Quantizer}, \ref{sec:Ablation_Study}, and \ref{sec:Experimental_Results} respectively.

\section{Hypergraph Convolution}
\label{sec:Hypergraph_Convolution}
\vspace*{-2mm}
\noindent{\textbf{Hypergraph:}} An undirected weighted graph $G=(V, E, W)$ is a structural representative of non-Euclidean data points in terms of nodes ($V$) and associated edges ($E$) along with their weights ($W$). Each edge in $G$ can only epitomize two nodes. The hypergraph \( \mathcal{H} = (\mathcal{V}, \mathcal{E}, \mathcal{W}) \) gives an excellent solution to achieve the semantic relation between more than two nodes. Here, $\mathcal{V}$ represents the nodes, same as in $G$, $\mathcal{E}$, and $\mathcal{W}$ denote the set of hyperedges and their corresponding weights, respectively. The hypergraph is represented by an incidence matrix \( \mathbf{H} \in \{0,1\}^{|\mathcal{V}| \times |\mathcal{E}|} \)
\vspace*{-2mm}
\begin{equation}
	\mathbf{H}_{i,j}=
	\begin{cases}
		1 & \text{   if node  $i$ belongs to hyperedge $j$ } \\
		0 & \text{   otherwise}
	\end{cases}
\vspace*{-2mm}
\end{equation}
A diagonal matrix $\mathbf{H_{w}} \in \mathbb{R}^{|\mathcal{E}| \times |\mathcal{E}|}$ is used to store the weights of hyperedges. The degree of a node $d_v(.)$ is the sum of the weights of all hyperedges associated with that node $v$, and the degree of a hyperedge $d_e(.)$ is the number of nodes contained in hyperedge $e$. $\mathbf{D_{v}}$ and $\mathbf{D_{e}}$ are diagonal degree matrices of nodes and hyperedges, respectively.

\noindent{\textbf{Hypergraph Convolution:}} A hypergraph convolution is defined as
\vspace*{-2mm}
\begin{equation}
	\mathbf{Y_{h}} = \mathbf{D_{v}}^{-1/2}\mathbf{H}\mathbf{H_{w}}\mathbf{{D_{e}}^{-1}}\mathbf{H^{T}}\mathbf{D_{v}}^{-1/2}\mathbf{X_1}\theta
	\vspace*{-2mm}
\end{equation}
where $\mathbf{X_{1}} \in \mathbb{R}^{|\mathcal{V}| \times C_{1}}$ is the input node feature matrix and $\theta \in \mathbb{R}^{C_{1} \times C_{2}}$ is the convolutional filter on hypergraph. We use the notation $HGConv({X}, \mathbf{H^n},\mathbf{{H_{w}}^n})$ to represent hypergraph convolution on the input feature  $X$. 

\section{Temporal attention in STA-HT block}
\label{sec:Temporal_attention}
\vspace*{-2mm}
In the STA-HT block, temporal attention is applied to the hyperedge features $H_f$. Attention weights for different timestamps are calculated and multiplied with $H_f$. We adopt a squeeze-excitation block to calculate the attention weights as in \cite{hu2019squeezeandexcitationnetworks}. A linear layer squeezes the number of temporal dimensions with ReLU activation, and another linear layer excites the temporal dimension to bring it back to its original count with sigmoid activation. These weights are multiplied to residual hyperedge features $H_f$ to get attentive features.

\section{Hypergraph Quantizer}
\label{sec:Hypergraph_Quantizer}
\vspace*{-2mm}
A codebook, which is formed by the trainable vectors \{$ c_{1},...., c_{k}$\}, is utilized to group the nodes that form input-dependent hyperedges. All these vectors can be represented by a tensor \( {C} \in \mathbb{R}^{ k  \times D} \). The intermediate output embedding \( {p} \in \mathbb{R}^{ V  \times d} \) for a single human skeleton from the second FAHT unit is passed through a decoder which is similar to the HypDec block in terms of architecture but comprising a different set of parameters. The $j^{th}$ node is assigned to hyperedge $ind_{j}$ in the following manner.
\vspace*{-2mm}
\begin{equation}
	ind_{j} = argmin_{i} \lVert c_{i}-p_{j} \rVert_2
	\vspace*{-2mm}
\end{equation}
Subsequently, we pass the incidence matrix ${{Hq}^n}$, representing the input-dependent hyperedge, to the next FAHT block. We optimize the distance between vector priors and node embeddings by allowing the vectors to be learnable, thereby enhancing adaptability. This approach closely resembles a clustering process. The mean squared error between the node embedding and $c_{j}$ is assigned as one of the loss functions for the model training.
\vspace*{-2mm}
\begin{equation}
	\mathcal{L}_{\text{quant}} = (1/V)\sum_{i=1}^{V} \lVert c_{ind_{i}}-sg(p_{i}) \rVert_2 ^{2}.
	\vspace*{-2mm}
\end{equation}
Here, sg denotes the gradient operator with respect to $p_{i}$, which is set as zero during backpropagation. 

To learn the weight of hyperedges, we use quantized embeddings, which are defined as 
\vspace*{-2mm}
\begin{equation}
	p_{quant_j}=c_{ind_{j}}.
	\vspace*{-2mm}
\end{equation}
\noindent We calculate ${{Hq_{w}}^n}$ after determining the weights of each hyperedge by advancing the quantized node embeddings through an MLP layer. Finally, ${{Hq_{w}}^n}$ is passed to the next set of FAHT blocks.

The $p_{quant_j}$ generation during the quantization process is a non-differentiable phenomenon, which is bypassed during backpropagation. We follow the straight-through estimator approach [refer to paper] for backpropagation which means the gradients from $p_{quant}$ are directly copied to $p$.

\section{Comprehensive Ablation Study}
\label{sec:Ablation_Study}
\vspace*{-2mm}
We evaluate our model using different values of related hyperparameters to get the best performance. These hyperparameters are the number of channels in transformer layers ($c$), the number of transformer blocks ($L$), and the number of hyperedges ($k$). Fig. \ref{fig:line_graph} shows the plots for model performance (accuracy in \%) vs number of epochs for the NTU RGB+D 60 dataset in the X-sub setting. As shown in this figure, we obtained the best results for $c=216$, $L=10$, and $k=5$. We also generate the t-SNE plots of the above hyperparameters, taking the same corresponding values as shown in Fig. \ref{fig:tsne} to consolidate our ablation. The best separation of different classes is visually evident for 10 transformer blocks with a channel count of 216 and hyperedges of 5.

\section{Experimental Results}
\label{sec:Experimental_Results}
\vspace*{-2mm}
Apart from the above, we have also analyzed the impact of various modules on individual class performances. As shown in Table \ref{tab:Ad-HGformer_quant1}, we have seen that the accuracies related to most of the individual classes are increasing due to the effectiveness of various modules in AutoregAd-HGformer.


\FloatBarrier
\begin{figure*}[h]
	\centering
	\captionsetup{justification=centering}
	\includegraphics[width=1.0\linewidth]{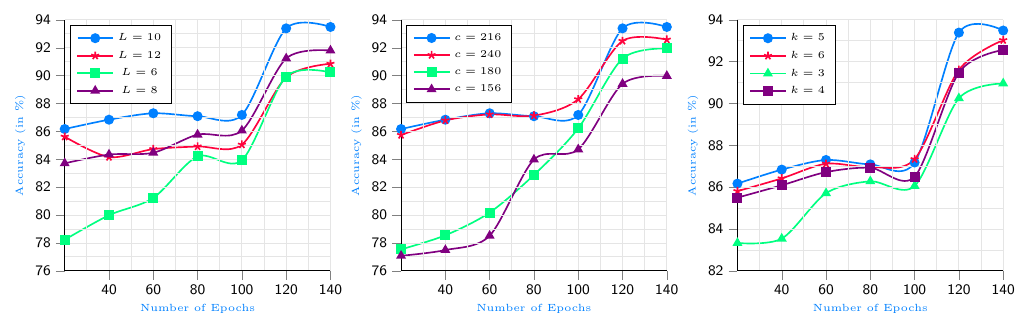}
	\caption{Epoch-wise performance (accuracy in \%) comparison of the proposed AutoregAd-HGformer for \textbf{Left:} transformer block counts ($L$), \textbf{Middle:} transformer channel counts ($c$), \textbf{Right:} hyperedge count ($k$). [on NTU RGB+D 60(X-sub)]}
	\label{fig:line_graph}
\end{figure*}

\begin{figure*}[h]
	\centering
	\captionsetup{justification=centering}
	\includegraphics[width=1.0\linewidth]{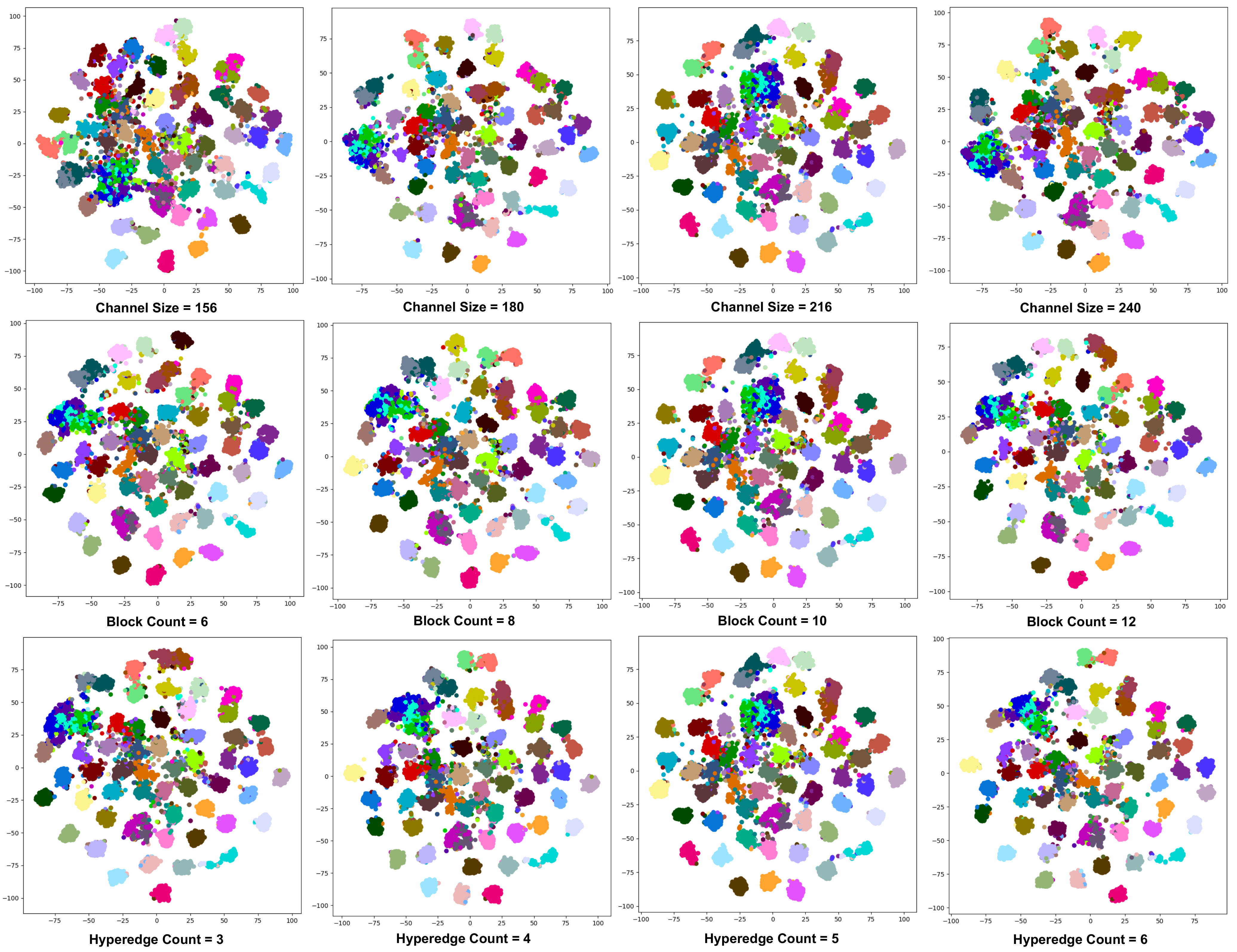}
	\caption{t-SNE \cite{van2008visualizing} plot of \textbf{Top:} transformer block counts ($L$), \textbf{Middle:} transformer channel counts ($c$), \textbf{Bottom:} hyperedge count ($k$). [on NTU RGB+D 60(X-sub)]}
	\label{fig:tsne}
\end{figure*}

\begin{table*}[h]
	\vspace*{-2mm}
	\begin{center}
		\captionsetup{justification=centering}
		\caption{Class-wise performance related to the impact of various modules in AutoregAd-HGformer on NTU RGB+D 60 X-Sub setting. OHG: Out-phase Hypergraph Generator, RL: Reconstruction Loss, THA: Temporal Hypergraph Attention, IHQ: In-phase Hyergraph Quantizer. The enhanced and reduced class-wise accuracy are given by \textcolor{purple}{$\uparrow$} and \textcolor{blue}{$\downarrow$} respectively.}
		\vspace*{-2mm}
		\scalebox{0.825}{
			\begin{tabular}{c|c|c|c|c|c}
				\multirow{1}{*}{\textbf{Methods}} & \multirow{1}{*}{\textbf{Baseline}} &  \multirow{1}{*}{\textbf{OHG}} & \multirow{1}{*}{\textbf{OHG+RL}} & \multirow{1}{*}{\textbf{OHG+RL+THA}} & \multirow{1}{*}{\textbf{OHG+RL+THA+IHQ}} \\
				\hline
				\backslashbox{\textbf{Action Labels}}{\textbf{Params(M)}} & 2.60 & 2.75 & 2.95 & 3.10 & 3.20  \\
				\hline
				\multirow{1}{*}{{drink water}} & \multirow{1}{*}{{87.70}} &  \multirow{1}{*}{{88.50 (\textcolor{purple}{0.8$\uparrow$})}} & \multirow{1}{*}{{88.52}} & \multirow{1}{*}{{88.25}} & \multirow{1}{*}{{88.95}} \\
				
				\hline
				\multirow{1}{*}{{eat meal/snack}} & \multirow{1}{*}{{78.89}} &  \multirow{1}{*}{{79.40(\textcolor{purple}{0.51$\uparrow$})}} & \multirow{1}{*}{{78.78}} & \multirow{1}{*}{{80.85}} & \multirow{1}{*}{{81.35}} \\
				
				\hline
				\multirow{1}{*}{{brushing teeth}} & \multirow{1}{*}{{91.68}} &  \multirow{1}{*}{{92.50(\textcolor{purple}{0.82$\uparrow$})}} & \multirow{1}{*}{{92.20}} & \multirow{1}{*}{{91.94}} & \multirow{1}{*}{{91.86}} \\
				
				\hline
				\multirow{1}{*}{{brushing hair}} & \multirow{1}{*}{{92.44}} &  \multirow{1}{*}{{92.92(\textcolor{purple}{0.48$\uparrow$})}} & \multirow{1}{*}{{93.95}} & \multirow{1}{*}{{95.12}} & \multirow{1}{*}{{94.26}} \\
				
				\hline
				\multirow{1}{*}{{drop}} & \multirow{1}{*}{{93.66}} &  \multirow{1}{*}{{93.94(\textcolor{purple}{0.28$\uparrow$})}} & \multirow{1}{*}{{94.85}} & \multirow{1}{*}{{95.56}} & \multirow{1}{*}{{94.78}} \\
				
				\hline
				\multirow{1}{*}{{pickup}} & \multirow{1}{*}{{98.38}} &  \multirow{1}{*}{{98.24(\textcolor{blue}{0.11$\downarrow$})}} & \multirow{1}{*}{{98.55}} & \multirow{1}{*}{{97.45}} & \multirow{1}{*}{{98.45}} \\
				
				\hline
				\multirow{1}{*}{{throw}} & \multirow{1}{*}{{94.48}} &  \multirow{1}{*}{{94.22(\textcolor{blue}{0.26$\downarrow$})}} & \multirow{1}{*}{{94.95}} & \multirow{1}{*}{{96.10}} & \multirow{1}{*}{{94.25}} \\
				\hline
				\multirow{1}{*}{{sitting down}} & \multirow{1}{*}{{98.88}} &  \multirow{1}{*}{{99.25(\textcolor{purple}{0.37$\uparrow$})}} & \multirow{1}{*}{{99.10}} & \multirow{1}{*}{{97.97}} & \multirow{1}{*}{{99.13}} \\
				\hline
				\multirow{1}{*}{{standing up (from sitting position}} & \multirow{1}{*}{{98.91}} &  \multirow{1}{*}{{99.24(\textcolor{purple}{0.33$\uparrow$})}} & \multirow{1}{*}{{98.90}} & \multirow{1}{*}{{99.15}} & \multirow{1}{*}{{99.27}} \\
				\hline
				\multirow{1}{*}{{clapping}} & \multirow{1}{*}{{86.86}} &  \multirow{1}{*}{{86.80(\textcolor{blue}{0.06$\downarrow$})}} & \multirow{1}{*}{{86.85}} & \multirow{1}{*}{{88.10}} & \multirow{1}{*}{{87.10}} \\
				\hline
				\multirow{1}{*}{{reading}} & \multirow{1}{*}{{61.79}} &  \multirow{1}{*}{{67.81(\textcolor{purple}{6.02$\uparrow$})}} & \multirow{1}{*}{{67.92}} & \multirow{1}{*}{{66.98}} & \multirow{1}{*}{{67.18}} \\
				\hline
				\multirow{1}{*}{{writing}} & \multirow{1}{*}{{71.63}} &  \multirow{1}{*}{{77.75(\textcolor{purple}{6.12$\uparrow$})}} & \multirow{1}{*}{{76.85}} & \multirow{1}{*}{{78.00}} & \multirow{1}{*}{{77.32}} \\
				\hline
				\multirow{1}{*}{{tear up paper}} & \multirow{1}{*}{{96.49}} &  \multirow{1}{*}{{96.63(\textcolor{purple}{0.14$\uparrow$})}} & \multirow{1}{*}{{96.60}} & \multirow{1}{*}{{96.06}} & \multirow{1}{*}{{96.57}} \\
				\hline
				\multirow{1}{*}{{wear jacket}} & \multirow{1}{*}{{98.58}} &  \multirow{1}{*}{{98.71(\textcolor{purple}{0.13$\uparrow$})}} & \multirow{1}{*}{{98.55}} & \multirow{1}{*}{{97.43}} & \multirow{1}{*}{{98.55}} \\
				\hline
				\multirow{1}{*}{{take off jacket}} & \multirow{1}{*}{{98.95}} &  \multirow{1}{*}{{99.17(\textcolor{purple}{0.22$\uparrow$})}} & \multirow{1}{*}{{99.25}} & \multirow{1}{*}{{99.48}} & \multirow{1}{*}{{99.18}} \\
				\hline
				\multirow{1}{*}{{wear a shoe}} & \multirow{1}{*}{{67.88}} &  \multirow{1}{*}{{70.01(\textcolor{purple}{2.13$\uparrow$})}} & \multirow{1}{*}{{71.74}} & \multirow{1}{*}{{73.16}} & \multirow{1}{*}{{73.78}} \\
				\hline
				\multirow{1}{*}{{take off a shoe}} & \multirow{1}{*}{{85.88}} &  \multirow{1}{*}{{86.10(\textcolor{purple}{0.22$\uparrow$})}} & \multirow{1}{*}{{86.37}} & \multirow{1}{*}{{85.00}} & \multirow{1}{*}{{84.00}} \\
				\hline
				\multirow{1}{*}{{wear on glasses}} & \multirow{1}{*}{{95.25}} &  \multirow{1}{*}{{94.80(\textcolor{blue}{0.45$\downarrow$})}} & \multirow{1}{*}{{93.67}} & \multirow{1}{*}{{95.95}} & \multirow{1}{*}{{95.15}} \\
				\hline
				\multirow{1}{*}{{take off glasses}} & \multirow{1}{*}{{96.46}} &  \multirow{1}{*}{{96.47(\textcolor{purple}{0.01$\uparrow$})}} & \multirow{1}{*}{{97.56}} & \multirow{1}{*}{{98.58}} & \multirow{1}{*}{{95.62}} \\
				\hline
				\multirow{1}{*}{{put on a hat/cap}} & \multirow{1}{*}{{98.42}} &  \multirow{1}{*}{{99.00(\textcolor{purple}{0.58$\uparrow$})}} & \multirow{1}{*}{{98.65}} & \multirow{1}{*}{{98.75}} & \multirow{1}{*}{{98.75}} \\
				\hline
				\multirow{1}{*}{{take off a hat/cap}} & \multirow{1}{*}{{98.95}} &  \multirow{1}{*}{{99.18(\textcolor{purple}{0.23$\uparrow$})}} & \multirow{1}{*}{{98.32}} & \multirow{1}{*}{{97.95}} & \multirow{1}{*}{{98.95}} \\
				\hline
				\multirow{1}{*}{{cheer up}} & \multirow{1}{*}{{94.83}} &  \multirow{1}{*}{{94.90(\textcolor{purple}{0.07$\uparrow$})}} & \multirow{1}{*}{{95.20}} & \multirow{1}{*}{{95.89}} & \multirow{1}{*}{{94.89}} \\
				\hline
				\multirow{1}{*}{{hand waving}} & \multirow{1}{*}{{96.26}} &  \multirow{1}{*}{{96.00(\textcolor{blue}{0.26$\downarrow$})}} & \multirow{1}{*}{{97.10}} & \multirow{1}{*}{{97.25}} & \multirow{1}{*}{{96.95}} \\
				\hline
				\multirow{1}{*}{{kicking something}} & \multirow{1}{*}{{97.76}} &  \multirow{1}{*}{{97.92(\textcolor{purple}{0.16$\uparrow$})}} & \multirow{1}{*}{{97.68}} & \multirow{1}{*}{{97.85}} & \multirow{1}{*}{{98.35}} \\
				\hline
				\multirow{1}{*}{{reach into pocket}} & \multirow{1}{*}{{87.55}} &  \multirow{1}{*}{{87.64(\textcolor{purple}{0.09$\uparrow$})}} & \multirow{1}{*}{{88.37}} & \multirow{1}{*}{{87.97}} & \multirow{1}{*}{{91.57}} \\
				\hline
				\multirow{1}{*}{{hopping (one foot jumping)}} & \multirow{1}{*}{{98.94}} &  \multirow{1}{*}{{99.26(\textcolor{purple}{0.32$\uparrow$})}} & \multirow{1}{*}{{99.11}} & \multirow{1}{*}{{98.91}} & \multirow{1}{*}{{98.79}} \\
				\hline
				\multirow{1}{*}{{jump up}} & \multirow{1}{*}{{99.25}} &  \multirow{1}{*}{{99.60(\textcolor{purple}{0.35$\uparrow$})}} & \multirow{1}{*}{{99.45}} & \multirow{1}{*}{{98.75}} & \multirow{1}{*}{{99.25}} \\
				\hline
				\multirow{1}{*}{{make a phone call/answer phone}} & \multirow{1}{*}{{93.13}} &  \multirow{1}{*}{{93.55(\textcolor{purple}{0.42$\uparrow$})}} & \multirow{1}{*}{{94.35}} & \multirow{1}{*}{{95.55}} & \multirow{1}{*}{{96.53}} \\
				\hline
				\multirow{1}{*}{{playing with phone/tablet}} & \multirow{1}{*}{{79.82}} &  \multirow{1}{*}{{83.60(\textcolor{purple}{3.78$\uparrow$})}} & \multirow{1}{*}{{81.38}} & \multirow{1}{*}{{81.47}} & \multirow{1}{*}{{82.56}} \\
				\hline
				\multirow{1}{*}{{typing on a keyboard}} & \multirow{1}{*}{{76.58}} &  \multirow{1}{*}{{81.17(\textcolor{purple}{4.59$\uparrow$})}} & \multirow{1}{*}{{83.27}} & \multirow{1}{*}{{84.66}} & \multirow{1}{*}{{84.86}} \\
				\hline
				\multirow{1}{*}{{pointing to something with finger}} & \multirow{1}{*}{{82.88}} &  \multirow{1}{*}{{83.24(\textcolor{purple}{0.36$\uparrow$})}} & \multirow{1}{*}{{83.95}} & \multirow{1}{*}{{84.34}} & \multirow{1}{*}{{85.49}} \\
				
				\hline
				\multirow{1}{*}{{taking a selfie}} & \multirow{1}{*}{{95.42}} &  \multirow{1}{*}{{95.80(\textcolor{purple}{0.38$\uparrow$})}} & \multirow{1}{*}{{96.06}} & \multirow{1}{*}{{96.66}} & \multirow{1}{*}{{97.82}} \\
				
				\hline
				\multirow{1}{*}{{check time (from watch)}} & \multirow{1}{*}{{94.32}} &  \multirow{1}{*}{{94.91(\textcolor{purple}{0.59$\uparrow$})}} & \multirow{1}{*}{{94.95}} & \multirow{1}{*}{{95.40}} & \multirow{1}{*}{{94.67}} \\
				
				\hline
				\multirow{1}{*}{{rub two hands together}} & \multirow{1}{*}{{92.91}} &  \multirow{1}{*}{{92.70(\textcolor{blue}{0.21$\downarrow$})}} & \multirow{1}{*}{{93.15}} & \multirow{1}{*}{{92.87}} & \multirow{1}{*}{{93.96}} \\
				
				\hline
				\multirow{1}{*}{{nod head/bow}} & \multirow{1}{*}{{98.85}} &  \multirow{1}{*}{{99.35(\textcolor{purple}{0.50$\uparrow$})}} & \multirow{1}{*}{{99.48}} & \multirow{1}{*}{{98.18}} & \multirow{1}{*}{{98.83}} \\
				
				\hline
				\multirow{1}{*}{{shake head}} & \multirow{1}{*}{{96.48}} &  \multirow{1}{*}{{97.60(\textcolor{purple}{1.12$\uparrow$})}} & \multirow{1}{*}{{98.35}} & \multirow{1}{*}{{99.02}} & \multirow{1}{*}{{98.61}} \\
				
				\hline
				\multirow{1}{*}{{wipe face}} & \multirow{1}{*}{{88.39}} &  \multirow{1}{*}{{89.70(\textcolor{purple}{1.31$\uparrow$})}} & \multirow{1}{*}{{89.89}} & \multirow{1}{*}{{90.39}} & \multirow{1}{*}{{91.76}} \\
				\hline
				\multirow{1}{*}{{salute}} & \multirow{1}{*}{{96.38}} &  \multirow{1}{*}{{96.30(\textcolor{blue}{0.08$\downarrow$})}} & \multirow{1}{*}{{96.47}} & \multirow{1}{*}{{97.05}} & \multirow{1}{*}{{96.89}} \\
				\hline
				\multirow{1}{*}{{put the palms together}} & \multirow{1}{*}{{98.86}} &  \multirow{1}{*}{{98.59(\textcolor{blue}{0.27$\downarrow$})}} & \multirow{1}{*}{{98.46}} & \multirow{1}{*}{{98.83}} & \multirow{1}{*}{{98.96}} \\
				\hline
				\multirow{1}{*}{{cross hands in front (say stop)}} & \multirow{1}{*}{{97.72}} &  \multirow{1}{*}{{98.35(\textcolor{purple}{0.63$\uparrow$})}} & \multirow{1}{*}{{99.15}} & \multirow{1}{*}{{98.15}} & \multirow{1}{*}{{98.94}} \\
				\hline
				\multirow{1}{*}{{sneeze/cough}} & \multirow{1}{*}{{85.24}} &  \multirow{1}{*}{{84.50(\textcolor{blue}{0.74$\downarrow$})}} & \multirow{1}{*}{{84.75}} & \multirow{1}{*}{{85.64}} & \multirow{1}{*}{{86.62}} \\
				\hline
				\multirow{1}{*}{{staggering}} & \multirow{1}{*}{{99.48}} &  \multirow{1}{*}{{99.63(\textcolor{purple}{0.15$\uparrow$})}} & \multirow{1}{*}{{99.48}} & \multirow{1}{*}{{98.88}} & \multirow{1}{*}{{99.26}} \\
				\hline
				\multirow{1}{*}{{falling}} & \multirow{1}{*}{{99.52}} &  \multirow{1}{*}{{99.87(\textcolor{purple}{0.35$\uparrow$})}} & \multirow{1}{*}{{99.82}} & \multirow{1}{*}{{99.48}} & \multirow{1}{*}{{99.36}} \\
				\hline
				\multirow{1}{*}{{touch head (headache)}} & \multirow{1}{*}{{88.98}} &  \multirow{1}{*}{{89.98(\textcolor{purple}{1.00$\uparrow$})}} & \multirow{1}{*}{{90.10}} & \multirow{1}{*}{{91.55}} & \multirow{1}{*}{{91.87}} \\
				\hline
				\multirow{1}{*}{{touch chest (stomachache/heart pain)}} & \multirow{1}{*}{{96.68}} &  \multirow{1}{*}{{96.35(\textcolor{blue}{0.33$\downarrow$})}} & \multirow{1}{*}{{96.53}} & \multirow{1}{*}{{96.35}} & \multirow{1}{*}{{97.78}} \\
				\hline
				\multirow{1}{*}{{touch back (backache)}} & \multirow{1}{*}{{97.49}} &  \multirow{1}{*}{{97.71(\textcolor{purple}{0.22$\uparrow$})}} & \multirow{1}{*}{{97.48}} & \multirow{1}{*}{{97.03}} & \multirow{1}{*}{{98.72}} \\
				\hline
				\multirow{1}{*}{{touch neck (neckache)}} & \multirow{1}{*}{{92.57}} &  \multirow{1}{*}{{92.22(\textcolor{blue}{0.35$\downarrow$})}} & \multirow{1}{*}{{93.67}} & \multirow{1}{*}{{93.25}} & \multirow{1}{*}{{95.44}} \\
				\hline
				\multirow{1}{*}{{nausea or vomiting condition}} & \multirow{1}{*}{{89.08}} &  \multirow{1}{*}{{89.40(\textcolor{purple}{0.32$\uparrow$})}} & \multirow{1}{*}{{89.22}} & \multirow{1}{*}{{89.87}} & \multirow{1}{*}{{90.74}} \\
				\hline
				\multirow{1}{*}{{use a fan/feeling warm}} & \multirow{1}{*}{{92.46}} &  \multirow{1}{*}{{92.02(\textcolor{blue}{0.44$\downarrow$})}} & \multirow{1}{*}{{93.19}} & \multirow{1}{*}{{93.18}} & \multirow{1}{*}{{94.26}} \\
				\hline
				\multirow{1}{*}{{punching/slapping other person}} & \multirow{1}{*}{{95.23}} &  \multirow{1}{*}{{95.99(\textcolor{purple}{0.76$\uparrow$})}} & \multirow{1}{*}{{95.72}} & \multirow{1}{*}{{94.99}} & \multirow{1}{*}{{95.63}} \\
				\hline
				\multirow{1}{*}{{kicking other person}} & \multirow{1}{*}{{97.37}} &  \multirow{1}{*}{{97.88(\textcolor{purple}{0.51$\uparrow$})}} & \multirow{1}{*}{{97.83}} & \multirow{1}{*}{{97.68}} & \multirow{1}{*}{{98.93}} \\
				\hline
				\multirow{1}{*}{{pushing other person}} & \multirow{1}{*}{{99.15}} &  \multirow{1}{*}{{99.60(\textcolor{purple}{0.45$\uparrow$})}} & \multirow{1}{*}{{99.37}} & \multirow{1}{*}{{98.83}} & \multirow{1}{*}{{99.42}} \\
				\hline
				\multirow{1}{*}{{pat on back of other person}} & \multirow{1}{*}{{95.71}} &  \multirow{1}{*}{{95.71(\textcolor{purple}{0.00$\uparrow$})}} & \multirow{1}{*}{{96.36}} & \multirow{1}{*}{{96.68}} & \multirow{1}{*}{{96.88}} \\
				\hline
				\multirow{1}{*}{point finger at the other person} & \multirow{1}{*}{{94.80}} &  \multirow{1}{*}{{94.92(\textcolor{purple}{0.12$\uparrow$})}} & \multirow{1}{*}{{96.00}} & \multirow{1}{*}{{96.45}} & \multirow{1}{*}{{96.18}} \\
				\hline
				\multirow{1}{*}{{hugging other person}} & \multirow{1}{*}{{99.55}} &  \multirow{1}{*}{{99.60(\textcolor{purple}{0.05$\uparrow$})}} & \multirow{1}{*}{{99.55}} & \multirow{1}{*}{{99.07}} & \multirow{1}{*}{{99.28}} \\
				\hline
				\multirow{1}{*}{{giving something to other person}} & \multirow{1}{*}{{97.55}} &  \multirow{1}{*}{{97.47(\textcolor{blue}{0.08$\downarrow$})}} & \multirow{1}{*}{{96.72}} & \multirow{1}{*}{{97.83}} & \multirow{1}{*}{{97.71}} \\
				\hline
				\multirow{1}{*}{{touch other person's+pocket}} & \multirow{1}{*}{{98.34}} &  \multirow{1}{*}{{98.90(\textcolor{purple}{0.56$\uparrow$})}} & \multirow{1}{*}{{97.62}} & \multirow{1}{*}{{98.75}} & \multirow{1}{*}{{98.91}} \\
				\hline
				\multirow{1}{*}{{handshaking}} & \multirow{1}{*}{{98.27}} &  \multirow{1}{*}{{98.63(\textcolor{purple}{0.36$\uparrow$})}} & \multirow{1}{*}{{98.78}} & \multirow{1}{*}{{97.36}} & \multirow{1}{*}{{99.07}} \\
				\hline
				\multirow{1}{*}{{walking towards each other}} & \multirow{1}{*}{{99.69}} &  \multirow{1}{*}{{99.66(\textcolor{blue}{0.03$\downarrow$})}} & \multirow{1}{*}{{99.54}} & \multirow{1}{*}{{98.86}} & \multirow{1}{*}{{99.41}} \\
				
				\hline
				\multirow{1}{*}{{walking apart from each other}} & \multirow{1}{*}{{98.35}} &  \multirow{1}{*}{{99.13(\textcolor{purple}{0.78$\uparrow$})}} & \multirow{1}{*}{{99.32}} & \multirow{1}{*}{{98.91}} & \multirow{1}{*}{{99.49}} \\
				
				\hline
				\multirow{1}{*}{\textbf{Overall}} & \multirow{1}{*}{{\textbf{92.90}}} &  \multirow{1}{*}{{\textbf{93.50}(\textcolor{purple}{0.60$\uparrow$})}} & \multirow{1}{*}{{\textbf{93.65}}} & \multirow{1}{*}{{\textbf{93.79}}} & \multirow{1}{*}{{\textbf{94.15}}} \\
				\bottomrule
				\bottomrule
		\end{tabular}}
		\label{tab:Ad-HGformer_quant1}
	\end{center}
\end{table*}
\FloatBarrier

\end{document}